\PassOptionsToPackage{table}{xcolor}
\documentclass{style}
\usepackage[T1]{fontenc}
\usepackage{amsmath,amssymb}
\usepackage{graphicx}
\usepackage{booktabs}
\usepackage{multirow}
\usepackage{array}
\usepackage{adjustbox}
\usepackage{listings}
\usepackage{enumitem}
\usepackage{fontawesome5}
\usepackage{wrapfig}
\usepackage{needspace}
\usepackage{microtype}
\usepackage[scaled=0.92]{zi4}
\makeatletter
\renewcommand{\BreakableUnderscore}{\leavevmode\prw@zbreak\textunderscore\discretionary{}{}{}\prw@zbreak}
\makeatother
\input{generated/values.tex}
\input{generated/v2_values.tex}
\makeatletter
\newcommand{\V}[1]{\ifcsname pv@#1\endcsname\csname pv@#1\endcsname\else
  \PackageError{values}{Unknown value #1}{Run scripts/paper_numbers.py}\fi}
\makeatother

\usepackage{makecell}
\definecolor{pendc}{HTML}{1F5FBF}
\newif\ifdraftmarks \draftmarksfalse   
\newcommand{\pend}[1]{\ifdraftmarks\textcolor{pendc}{#1}\else#1\fi}

\definecolor{toolink}{HTML}{596675}
\definecolor{codec}{HTML}{8C0000}
\definecolor{codeink}{HTML}{8C0000}
\colorlet{gaintint}{codec!6}

\newcommand{\gain}[1]{\cellcolor{gaintint}#1}
\newcommand{\hdtool}{\textcolor{toolink}{\textit{Tool}}}
\newcommand{\hdcode}{\textcolor{codeink}{\textit{Code}}}

\usepackage{pifont}
\newcommand{\abyes}{\checkmark}
\newcommand{\abno}{\textcolor{black!45}{\ding{55}}}
\newcommand{\abna}{\textcolor{black!45}{--}}
\definecolor{codebg}{HTML}{F7F6F2}
\definecolor{coderule}{HTML}{D9DCD2}
\definecolor{codeorange}{HTML}{8C0000}
\definecolor{codestr}{HTML}{3F6B3F}
\lstdefinestyle{py}{
  language=Python, basicstyle=\ttfamily\scriptsize, upquote=true, keywordstyle=\bfseries, stringstyle=\color{codestr},
  commentstyle=\color{gray}, showstringspaces=false, columns=fullflexible, keepspaces=true,
  breaklines=true, backgroundcolor=\color{codebg}, frame=single, rulecolor=\color{coderule}, framesep=3pt,
  xleftmargin=3pt, xrightmargin=3pt, aboveskip=3pt, belowskip=0pt,
  emph={robo}, emphstyle=\color{codeorange}\bfseries, numbers=none}
\setlist{leftmargin=*,itemsep=1pt,topsep=2pt,parsep=0pt}

\usepackage{xcolor}

\title{Fewer Tokens, Better Action:\\ GPT-6 Astra Robot Agents with \V{POOLED_SR_REL} Higher Success Rate
  but \V{POOLED_TOKENS_FEWER} Fewer Tokens}

\renewcommand{\authorfont}{\fontsize{11.5}{14}\selectfont}
\author{
  Ruiyang~Si\texorpdfstring{$^{1}$}{} \hspace{0.1cm}
  Jianxin~Bi\texorpdfstring{$^{1,2}$}{} \hspace{0.1cm}
  Shunyu~Yang\texorpdfstring{$^{1}$}{} \hspace{0.1cm}
  Rui~Ni\texorpdfstring{$^{1}$}{} \hspace{0.1cm}
  Wenbo~Huang\texorpdfstring{$^{1}$}{}\\
  Qiang~Wang\texorpdfstring{$^{1}$}{} \hspace{0.1cm}
  Shulong~Jiang\texorpdfstring{$^{1}$}{} \hspace{0.1cm}
  Yintianrun~Zhang\texorpdfstring{$^{1}$}{} \hspace{0.1cm}
  Feixiang~Ruan\texorpdfstring{$^{1}$}{} \hspace{0.1cm}
  Kewei~Zhang\texorpdfstring{$^{1}$}{}\\
  Duomin~Wang\texorpdfstring{$^{3}$}{} \hspace{0.1cm}
  Xiuyu~Li\texorpdfstring{$^{4}$}{} \hspace{0.1cm}
  Haiwen~Feng\texorpdfstring{$^{4}$}{} \hspace{0.1cm}
  Zhen~Dong\texorpdfstring{$^{3}$}{} \hspace{0.1cm}
  Daquan~Zhou\texorpdfstring{$^{1}$}{}
}

\affiliation[1]{Peking University}
\affiliation[2]{National University of Singapore}
\affiliation[3]{NVIDIA}
\affiliation[4]{Impossible Research}

\checkdata[
\raisebox{-0.2em}{\includegraphics[width=0.025\linewidth]{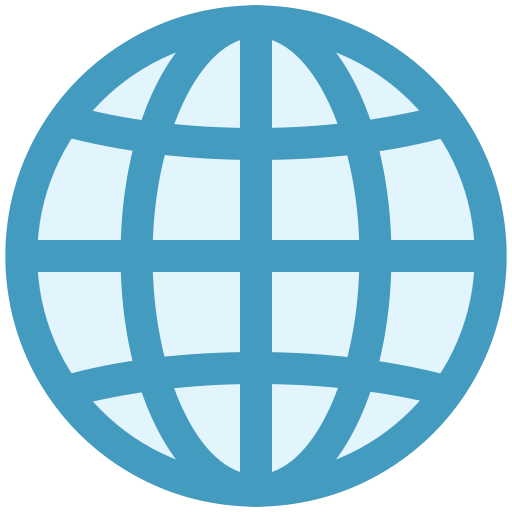}}~~Project Page]{\href{https://dagroup-pku.github.io/PyRUA-Lean/}{\texttt{https://dagroup-pku.github.io/PyRUA-Lean/}}
\\[-1.5ex]}

\checkdata[
\raisebox{-0.2em}{\includegraphics[width=0.025\linewidth]{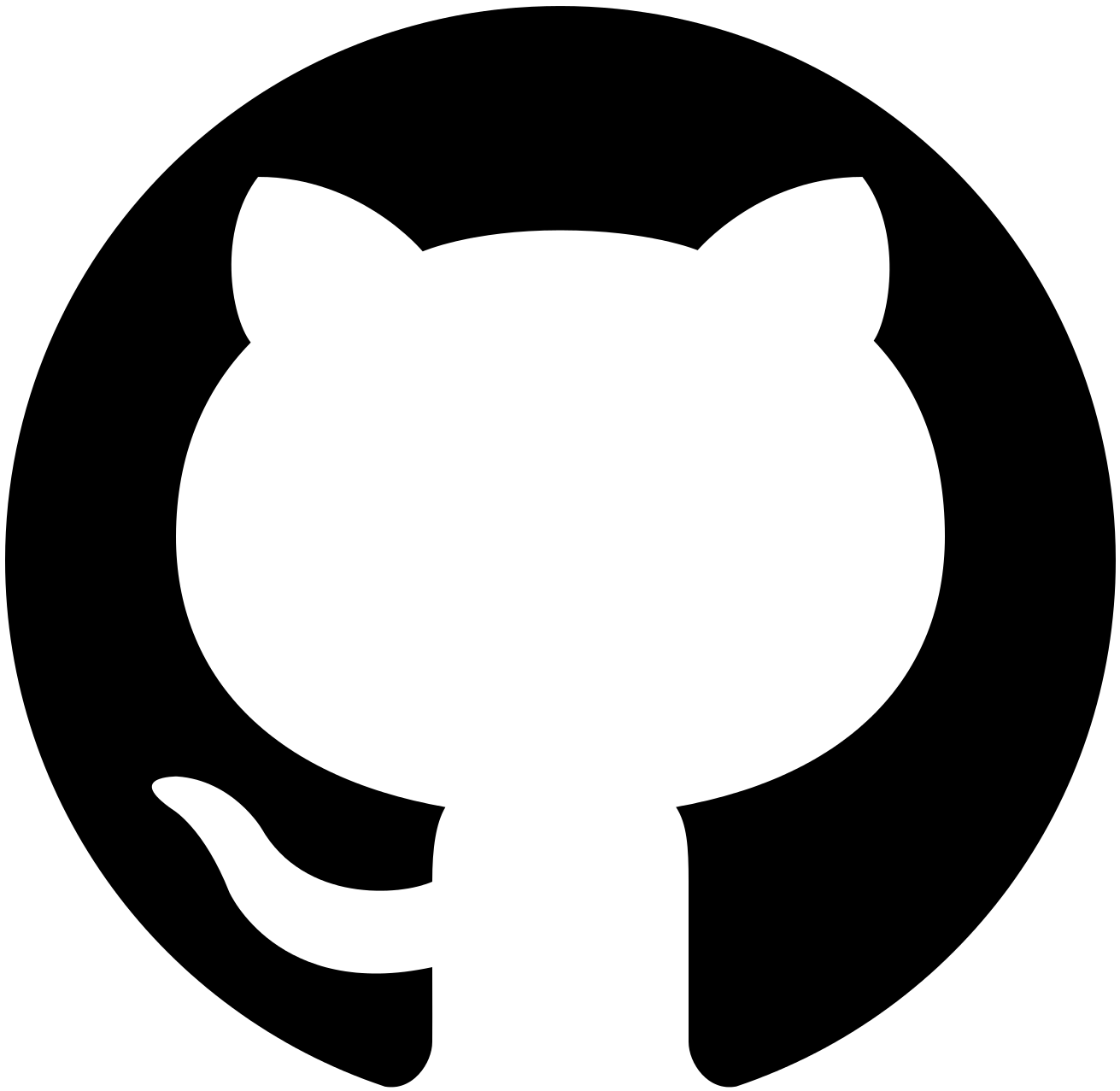}}~~GitHub Repo]{\href{https://github.com/DAGroup-PKU/PyRUA-Lean}{\texttt{https://github.com/DAGroup-PKU/PyRUA-Lean}}
\\[-1.7ex]}

\checkdata[
\raisebox{-0.2em}{\includegraphics[width=0.025\linewidth]{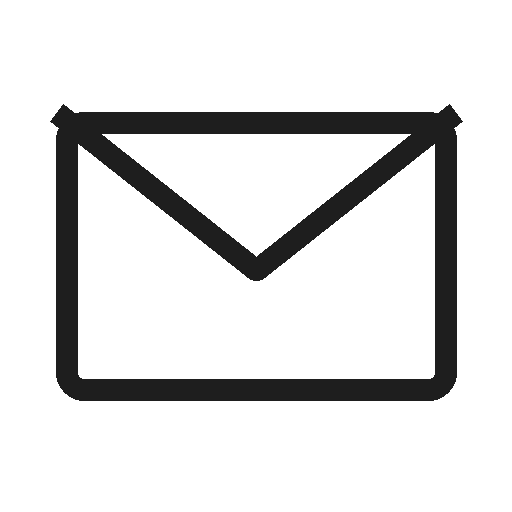}}~~Corresponding Author]{\href{mailto:zhoudaquan21@gmail.com}{\texttt{Daquan Zhou}}
\\[-1.7ex]}

\abstract{
Vision-language model (VLM) agents can control robots through visual feedback and action primitives, but repeated LLM calls and redundant observations incur substantial token overhead. We introduce PyRUA-Lean, an interactive code-execution framework in which the agent composes robot primitives, including vision-language-action (VLA) policies, into Python cells that check and retry locally and return only the images and state they request. Across 700 simulated task instances from LIBERO-PRO, RoboTwin 2.0, and RoboCasa365, it raises success from 63.1\% to 71.7\% over tool calling with the same GPT-6 Astra planner and primitives, using 49\% fewer LLM calls and 65\% fewer input tokens on jointly solved instances. On RoboDojo, without any VLA policy or privileged information, it reaches \V{RD_SR}\% average success rate (Score \V{RD_SCORE}), \V{RD_VS_TOP_SR} points above the top entry of the public leaderboard. Based on the PyRUA-Lean framework, we train PyRUA-Lean Coder v0.1-4B, an open 4B model, with GRPO on 45 LIBERO-PRO instances; it improves from \V{CD_B4_UNSEEN}\% to \V{CD_C4_UNSEEN}\% on unseen task instances, surpassing the Qwen3.5-27B base model.
}

\begin{document}
\maketitle

\ifdraftmarks
\par\nointerlineskip\vbox to 0pt{\vss\noindent{\small\pend{\textbf{Draft of v2.} Text and numbers in this colour are provisional (runs in progress); the last page lists them.}}\par\vskip -5pt}\nointerlineskip
\fi

\suppressfloats[t]
\begin{figure}[t!]
  \centering
  \includegraphics[width=\linewidth]{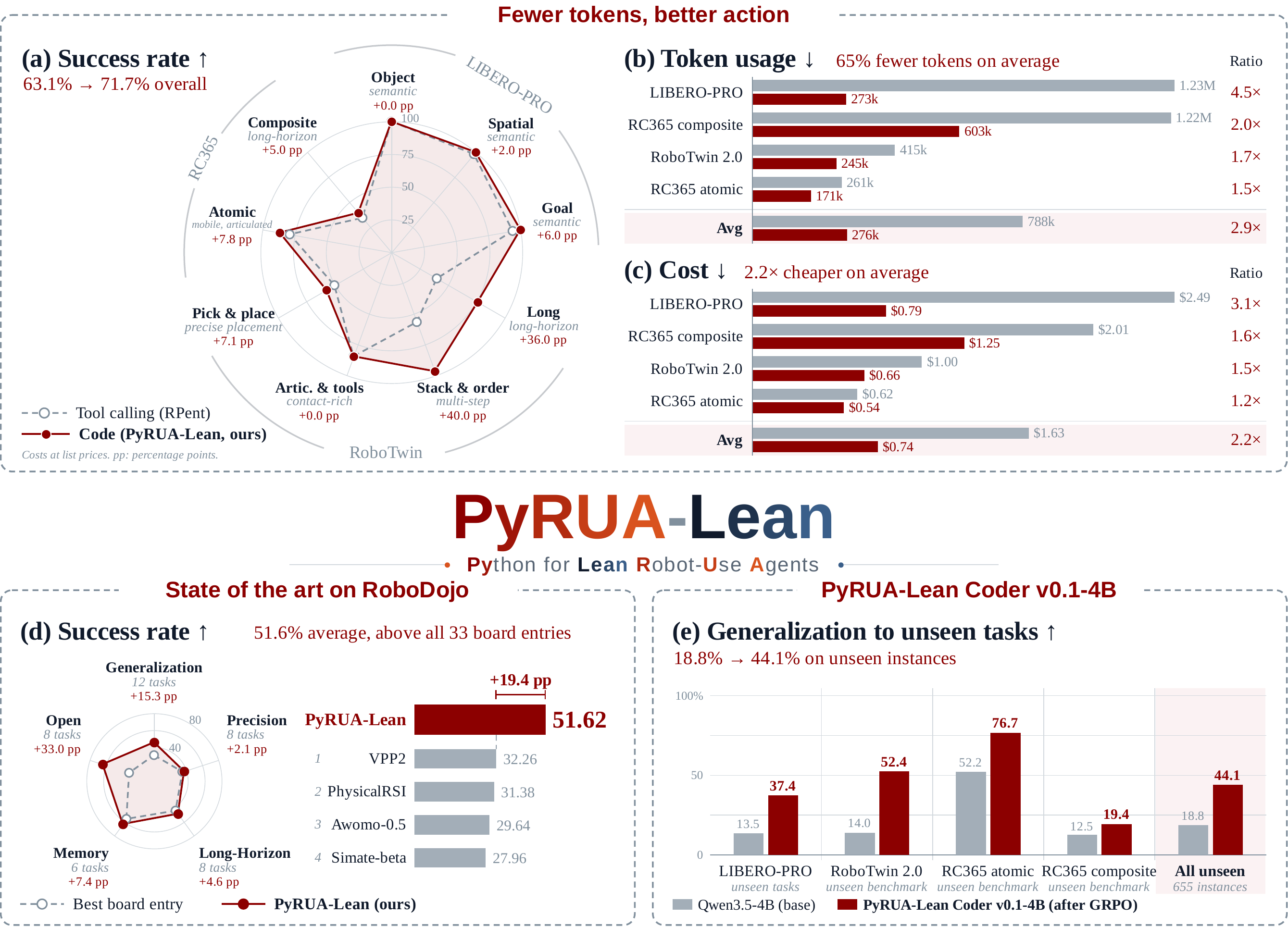}
  \caption{\textbf{PyRUA-Lean at a glance.} \emph{Top, the interface against tool calling} with the same GPT-6 Astra
    planner and primitives: \emph{(a)} success rate on all 700 instances, tagged by task property (semantic: perturbed
    layouts, objects, goals), RoboTwin 2.0 split by skill from task names; \emph{(b, c)} input tokens and GPT-6 Astra
    list-price dollars per solved episode, on instances both agents solved (Avg: mean over all of them).
    \emph{Bottom left, RoboDojo, (d):} success rate per capability dimension against the best entry of the public
    leaderboard in each dimension, and the average success rate of PyRUA-Lean and of the four highest-ranked of the
    board's \V{RD_BOARD_N} entries.
    \emph{Bottom right, PyRUA-Lean Coder v0.1-4B, (e):} generalization. Qwen3.5-4B is trained with GRPO on 45
    LIBERO-PRO task instances only; shown is its success rate on tasks and benchmarks it never saw, before (gray)
    and after training (red). All unseen pools these 655 instances. RC365: RoboCasa365.\V{CD_STAR_NOTE}\V{RD_STAR_NOTE}}
  \label{fig:overview}
\end{figure}

\section{Introduction}
\label{sec:intro}

Vision-language models (VLMs) can perform robot manipulation by interpreting visual observations and robot state, then invoking motion, perception, and grasping primitives, including vision-language-action (VLA) policies~\cite{ahn2022saycan,huang2022inner,zhang2026harness,su2026astra}. Their efficiency depends partly on how these primitives are exposed to the planner. In a tool-calling interface, an operation that depends on a previous tool's result generally requires another LLM call. Intermediate results also accumulate in the conversation and are processed again in every later LLM call; in systems such as RPent~\cite{zhang2026harness}, motion primitives automatically return multiple camera images at the end of each step of execution. Tasks requiring repeated localization, motion, and recovery can therefore incur substantial inference overhead, even when individual primitives are reliable.

Executable code can reduce this overhead. A program can compose primitives, inspect their outputs, and execute conditional branches or retries before returning control to the LLM. Code-based robot control is well established: Code as Policies and ProgPrompt generate robot programs from language instructions~\cite{liang2023code,singh2023progprompt}, while more recent systems incorporate execution feedback or refine programs across trials~\cite{fu2026capx,lu2026aspire}. Outside robotics, code-based agent interfaces have shown benefits in task completion and context efficiency~\cite{wang2024codeact,anthropic2025mcp,varda2025codemode}. These findings motivate our central question: \textit{how can an LLM agent compose robot primitives to reliably complete tasks while reducing token consumption through selective observation and context management?}

We introduce \textbf{PyRUA-Lean} (Python for Lean Robot-Use Agents),
an interactive code-execution interface for feedback-driven
primitive composition and selective observation. The agent generates Python programs that compose robot primitives, check intermediate outcomes, and condition subsequent actions on execution feedback. Intermediate results remain available in a persistent program state, while only explicitly printed outputs and requested camera images are returned to the LLM. This allows the agent to handle intermediate execution steps without additional LLM calls and to control which observations enter its context.

This report studies the interface from three sides (Figure~\ref{fig:overview}).
\emph{First}, we compare it with tool calling under controlled conditions. PyRUA-Lean builds on RPent's robot stacks and primitive implementations~\cite{zhang2026harness}, and we compare it with RPent's tool-calling agent under the same GPT-6 Astra planner, frozen VLA policies, evaluation task instances, and LLM-call budgets, on 700 simulated task instances from LIBERO-PRO, RoboTwin 2.0, and RoboCasa365. Both agents operate without cross-episode memory, and the comparison evaluates the interfaces as a whole, including their support for control flow, persistent program state, and observation delivery.
\emph{Second}, we ask how far the interface carries a frozen LLM on a benchmark built for pretrained robot policies. On RoboDojo~\cite{chen2026robodojo}, we give the same planner a small library of geometric primitives that we wrote ourselves, and follow the benchmark's evaluation protocol. The agent uses no VLA policy, no demonstrations, no fine-tuning, and no privileged information: everything it observes is the RGB observation that the benchmark sends to every policy.
\emph{Third}, we ask whether a small open model can learn the interface. Based on the PyRUA-Lean framework, we train Qwen3.5-4B~\cite{qwen2026qwen35} to write the cells itself, using GRPO~\cite{shao2024deepseekmath} with task success as the only reward, and call the resulting planner PyRUA-Lean Coder v0.1-4B.

Our contributions can be summarized as:
\begin{itemize}
    \item \textbf{PyRUA-Lean, a token-efficient code interface for robot agents.}
    The agent composes robot primitives, VLA policies included, into Python cells that check outcomes, keep state, and return only the observations they request. With the same planner and primitives as tool calling, it raises success from 63.1\% to 71.7\% and, on jointly solved instances, uses 49\% fewer LLM calls and 65\% fewer input tokens (Section~\ref{sec:framework}).

    \item \textbf{State-of-the-art results on RoboDojo.}
    With GPT-6 Astra as the planner, no learned policy acting on the robot, and no privileged information, PyRUA-Lean reaches \V{RD_SR}\% average success rate (Score \V{RD_SCORE}), \V{RD_VS_TOP_SR} points above the top entry of the public leaderboard and \V{RD_VS_ASTRA_X} times the same LLM controlling the end-effector directly, and is the highest on \V{RD_DIMS_BEST_PHRASE} capability dimensions (Section~\ref{sec:robodojo}).

    \item \textbf{PyRUA-Lean Coder v0.1-4B, an open planner trained to use the interface.}
    Using the PyRUA-Lean framework, we train Qwen3.5-4B with GRPO on only 45 LIBERO-PRO instances. On tasks and benchmarks it never saw, its success rate rises from \V{CD_B4_UNSEEN}\% to \V{CD_C4_UNSEEN}\%, above the Qwen3.5-27B base model; on RoboTwin 2.0 alone, from \V{CD_B4_RT}\% to \V{CD_C4_RT}\%. We will release the weights (Section~\ref{sec:coder}).
\end{itemize}

\section{Related Work}
\label{sec:related}

\textbf{LLM-based robot control.}
Language models have been used to select robot skills, interpret execution feedback, and generate motion commands. SayCan grounds skill selection in language instructions and learned affordances~\cite{ahn2022saycan}, while Inner Monologue incorporates environment feedback into planning~\cite{huang2022inner}. More recent systems support finer-grained control: Show-Harness exposes discrete semantic actions to a VLM~\cite{chen2026showharness}, and GPT-6 Astra has been evaluated as an embodied policy that generates or corrects robot actions~\cite{su2026astra}. VLA models provide learned visuomotor capabilities for language-conditioned control~\cite{brohan2023rt2,kim2024openvla,black2025pi0,intelligence2025pi05}. Harness VLA integrates frozen VLA policies with analytic primitives through RPent, a memory-guided agent framework~\cite{zhang2026harness}. PyRUA-Lean builds on RPent's robot stacks and primitive implementations, exposing these capabilities through an interactive Python interface. We compare this interface with RPent's tool-calling interface using the same planner and underlying primitives, with cross-episode memory disabled in both agents.

\textbf{Code-based robot control.}
Code as Policies and ProgPrompt generate executable robot programs from language instructions~\cite{liang2023code,singh2023progprompt}, while VoxPoser generates code that constructs spatial value maps for motion planning~\cite{huang2023voxposer}. Subsequent work explores execution feedback and program adaptation. CaP-X benchmarks coding agents across primitive abstraction levels and single- or multi-turn interaction modes~\cite{fu2026capx}, and VLCP periodically regenerates a control function from updated observations within an episode~\cite{naouali2026vlcp}. Other systems accumulate reusable capabilities across trials: Voyager develops a code-based skill library in Minecraft~\cite{wang2023voyager}, ASPIRE refines robot programs and consolidates experience into reusable skills~\cite{lu2026aspire}, and RoboRSI converts established tool-use workflows into executable routines as part of a robot self-improvement system~\cite{noematrix2026roborsi}. These studies establish program synthesis, feedback-driven execution, and skill reuse as complementary approaches to embodied control. Our focus is the success and inference cost of interactive code execution relative to tool calling over the same robot primitives, including VLA policies. We evaluate this comparison without cross-episode skill accumulation, while allowing state and code reuse within each episode.

\looseness=-1
\textbf{Code-based interfaces for efficient agents.}
Research on general-purpose agents provides a broader motivation for this comparison. ReAct interleaves reasoning with environment interaction~\cite{yao2023react}, and Toolformer studies the use of external APIs by language models~\cite{schick2023toolformer}. CodeAct shows that executable Python can improve agent performance relative to text or JSON action interfaces~\cite{wang2024codeact}, while SWE-agent demonstrates the importance of interface design for software-engineering agents~\cite{yang2024sweagent}. More recent systems use code to compose tool calls and process intermediate results outside the model context, reducing the information returned to the LLM~\cite{anthropic2025mcp,varda2025codemode}. PyRUA-Lean applies these interface principles to robot manipulation, where observations are visual and primitive execution may fail. We measure task success together with LLM-call usage, input-token consumption, and inference cost, and analyze how programmatic composition and observation passing contribute to the gains.

\section{PyRUA-Lean: Primitive Composition with Selective Observation}
\label{sec:method}
\label{sec:framework}

PyRUA-Lean is an interactive code-execution interface for LLM-based robot agents. It supports two complementary mechanisms: composing robot primitives into programs that respond to execution feedback, and controlling which intermediate results and visual observations enter the LLM context. The interface builds on RPent's robot stacks and primitive implementations~\cite{zhang2026harness}, including classical motion and perception primitives and learned VLA policies.

\begin{figure}[t]
  \centering
  \includegraphics[width=0.98\linewidth]{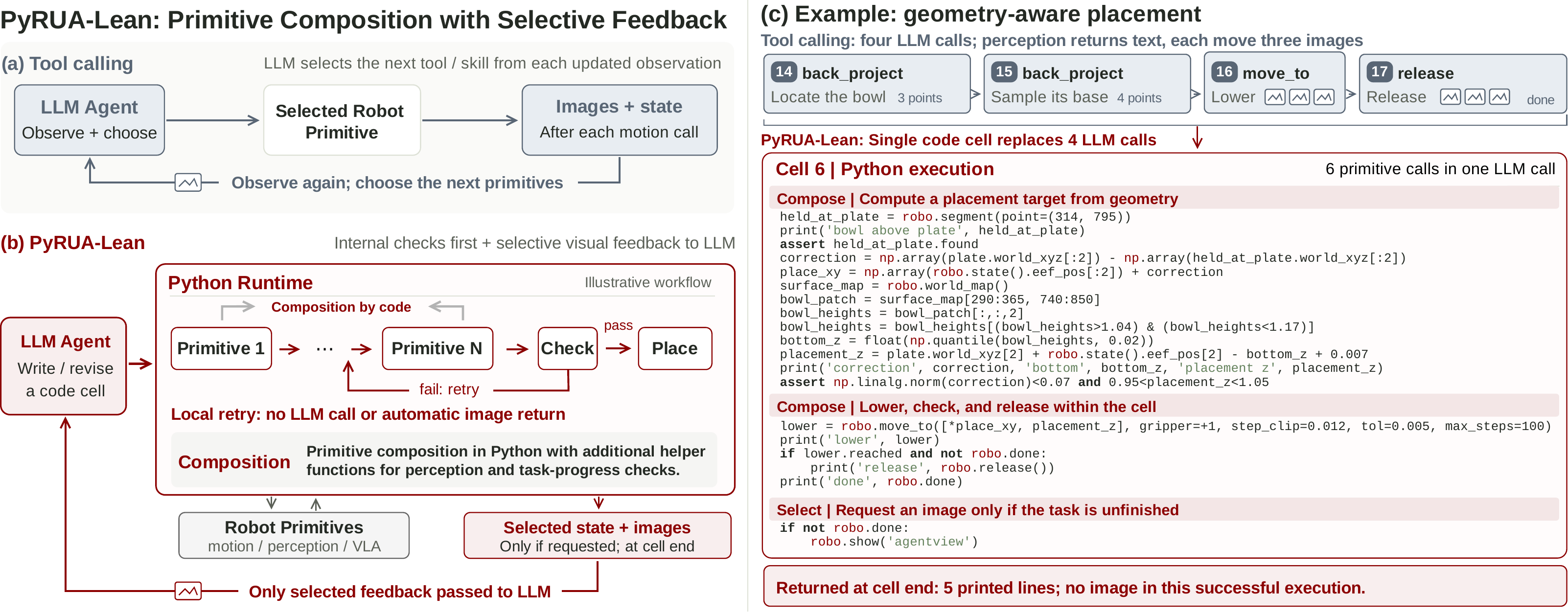}
    \caption{\textbf{PyRUA-Lean combines primitive composition with selective feedback.} (a) The tool-calling agent invokes robot primitives through tool calls. Operations that depend on preceding results generally require another LLM call, and motion calls automatically return images and state. (b) PyRUA-Lean instead generates Python cells that compose robot primitives with helper functions, conditional checks, and local retries. Intermediate execution stays within the runtime, while explicitly requested images and state messages are recorded and returned at cell end for replanning. (c) A recorded placement example illustrates both mechanisms: one PyRUA-Lean cell replaces the baseline's LLM calls 14--17, reducing four LLM calls to one. The cell computes a placement target from scene geometry (compose), conditionally executes lowering and release (compose), and requests an image only if the task remains unfinished (select). This successful execution returns five printed lines and no image, demonstrating how programmatic composition and selective feedback reduce repeated LLM calls. Appendix~\ref{app:calls} follows the whole episode.}
  \label{fig:method}
\end{figure}

\subsection{Robot Interface and Execution Model}
\label{sec:execution_model}

The robot is exposed as a Python object, \texttt{robo}, whose methods invoke primitives and return structured results, such as object positions and motion outcomes. The agent receives an API reference (Table~\ref{tab:interface} lists the primitives) and interacts with the robot through a \texttt{python(code)} tool. In an LLM call, the model generates a code cell; the runtime executes it, and the returned feedback informs the next call. GPT-6 Astra writes one cell per LLM call, but a call may hold several (Section~\ref{sec:coder_results}). A persistent Python namespace retains variables and helper functions across cells within an episode. Our evaluation uses no cross-episode memory.

An uncaught exception terminates the current cell and returns its traceback to the LLM. Variables assigned before the exception remain in the persistent namespace, allowing the agent to revise the failed operation in a subsequent cell. Before each primitive that acts on the robot, the runtime terminates the cell if the task has already succeeded. An external watchdog enforces the two-hour episode limit by interrupting any active primitive and stopping the planner. We impose no separate per-cell execution limit; in our setup, the Codex CLI returns a tool-call timeout after 300~s without cancelling the cell, and subsequent cells are queued.

\subsection{Feedback-Driven Primitive Composition}
\label{sec:primitive_composition}

Each cell can compose multiple primitives and condition subsequent operations on their outcomes. For example, a program may approach an object, attempt a grasp only if the approach succeeds, and retry or stop based on the resulting state. These checks and branches execute within the cell without additional LLM calls. At the end of the cell, the returned feedback allows the LLM to revise subsequent actions, combining program-level feedback with LLM-level replanning.

Beyond composing primitive calls, programs can \emph{compose} task-specific computations and control routines from existing observations and primitives, as illustrated by the geometric computation in Figure~\ref{fig:method}(c). For example, the agent computes the bowl's base height from a world-coordinate map to determine a placement target that no individual primitive directly returns. It can also define helper functions for repeated operations, such as a guarded descent. Computed values and helper functions remain available to later cells through the persistent namespace.

\subsection{Selective Observation and Context Management}
\label{sec:context_management}

Primitive results remain available in the runtime without being individually added to the LLM conversation. The agent specifies which feedback to return through explicit output statements and camera requests such as \texttt{robo.show}. Requested images and state messages are recorded at the points specified in the code and returned together at the end of the cell. 
In the episode summarized in Appendix~\ref{app:calls}, cell 5 requests a camera image after transporting the bowl. The placement cell shown in Figure~\ref{fig:method}(c) requests an image only if the task remains unfinished and therefore returns no images after successful placement.

Programmatic composition reduces the LLM calls needed for dependent operations, while selective feedback limits the outputs and images added to subsequent prompts. Persistent program state retains intermediate data outside the conversation. These mechanisms control new context rather than removing existing history: both agents run in the Codex CLI~\cite{openai2025codexcli}, which keeps the full conversation and compacts it, summarizing its earlier part, only when it approaches 272k tokens.

\begin{table}[t]
  \centering
  \caption{The two agents' interfaces: each primitive is an RPent tool for tool calling and a \texttt{robo} method with
    the same name and arguments for code, both running RPent's implementation. Names pooled over the benchmarks.}
  \label{tab:interface}
  \footnotesize
  \setlength{\tabcolsep}{5pt}
  \begin{tabular}{@{}lcc@{}}
    \toprule
    & \hdtool\ (RPent) & \hdcode\ (PyRUA-Lean) \\
    \midrule
    \multicolumn{3}{@{}l}{\emph{Primitives, the same in both}} \\
    Arm motion & \multicolumn{2}{c}{\texttt{move\_to}, \texttt{move\_pose}, \texttt{move\_delta}, \texttt{rotate\_wrist},
      \texttt{rotate\_pitch}} \\
    Gripper & \multicolumn{2}{c}{\texttt{set\_gripper}, \texttt{release}, \texttt{scripted\_grasp}} \\
    VLA policy & \multicolumn{2}{c}{\texttt{pi0\_pick}, \texttt{pi0\_doubled}, \texttt{lingbot\_act}, \texttt{rldx\_skill},
      \texttt{rldx\_arm}} \\
    Mobile base & \multicolumn{2}{c}{\texttt{navigate\_to}, \texttt{move\_base}} \\
    Perception & \multicolumn{2}{c}{\texttt{segment}, \texttt{back\_project}, \texttt{back\_project\_batch},
      \texttt{sample\_world\_xyz}, \texttt{query\_world\_map}} \\
    \midrule
    \multicolumn{3}{@{}l}{\emph{Observations}} \\
    State, success flag & with every move, or \texttt{view\_env\_state} & \texttt{robo.state()}, \texttt{robo.done} \\
    Camera images & three with every move & when the code calls \texttt{robo.show} \\
    World coordinates & through the perception tools & the same, or \texttt{robo.world\_map()} as one array \\
    \bottomrule
  \end{tabular}
\end{table}

\subsection{Evaluation Against Tool Calling}
\label{sec:experiments}

Our central question is whether an LLM agent can compose robot primitives to improve task completion while reducing token overhead through programmatic execution and selective feedback. We hold the planner, the primitives, and the task instances fixed, and evaluate PyRUA-Lean against a tool-calling baseline through three hypotheses (\textbf{H}):

\begin{itemize}
    \item \textbf{H1:}
    PyRUA-Lean achieves higher task success than tool calling under equal LLM-call budgets.

    \item \textbf{H2:} PyRUA-Lean requires fewer LLM calls and input tokens to solve tasks.

    \item \textbf{H3:}
    PyRUA-Lean retains its token efficiency when VLA policies or operating guides are absent.
\end{itemize}

\label{sec:experimental_setup}%

\textbf{Benchmarks.}
We evaluate 40 tasks from LIBERO-PRO's four perturbed suites~\cite{zhou2025liberopro,liu2023libero}, 50 dual-arm tasks from RoboTwin 2.0~\cite{chen2025robotwin2}, and 50 tasks from RoboCasa365's Target50 set~\cite{nasiriany2026robocasa365}. Each task is evaluated at five environment seeds, with one episode per agent for each task–seed pair, yielding 700 paired task instances and 1,400 episodes. More implementation details can be found in Appendix~\ref{app:benchmarks}.

\textbf{Agents.}
Both agents use GPT-6 Astra~\cite{openai2026astra} at high reasoning effort through the Codex CLI, with the same robot stacks, primitive implementations, and frozen VLA policies: $\pi_{0.5}$ on LIBERO-PRO~\cite{intelligence2025pi05}, LingBot-VLA on RoboTwin 2.0~\cite{wu2026pragmatic}, and RLDX-1 on RoboCasa365~\cite{kim2026rldx}. Both use SAM~3 for segmentation~\cite{carion2025sam3}. The baseline invokes RPent's primitives through tool calls; PyRUA-Lean composes them through its Python interface. Both receive the same operating guides where available; RoboCasa365 has none. Both agents operate without cross-episode memory.

\enlargethispage{\baselineskip}%
\textbf{Metrics.}
Success rate is measured over all task instances using each benchmark's success check, with budgets of 40 LLM calls per episode, or 100 for RoboCasa365 composite tasks, and two hours. These budgets limit LLM calls, not the number of primitive executions. We compare LLM calls, cumulative input tokens, and estimated monetary cost on instances solved by both agents, reporting whole-episode means and ratios of aggregate totals. Counts include post-success LLM calls and cached input tokens.
Cost-accounting details and all-episode costs are reported in
Appendix~\ref{app:cost}.

\begin{table}[t]
  \centering
  \caption{Main results. Success rate on all task instances; LLM calls, input tokens and dollars at list prices
    per solved episode, on the task instances both agents solved (means). \emph{Tool}: tool calling (RPent);
    \emph{Code}: PyRUA-Lean (ours). Shaded: code's gain, in points ($\Delta$) or how many times fewer ($\times$). \emph{All}
    pools the four benchmark groups, where LIBERO-PRO's long episodes dominate the token factor. RC365: RoboCasa365.}
  \label{tab:main}
  \footnotesize
  \setlength{\tabcolsep}{3.3pt}
  \begin{adjustbox}{max width=\linewidth}
  \begin{tabular}{@{}l|ccc|ccc|ccc|ccc@{}}
\toprule
 & \multicolumn{3}{c|}{All task instances} & \multicolumn{9}{c}{Per solved episode, on the task instances both agents solved} \\
\cmidrule(lr){2-4}\cmidrule(l){5-13}
 & \multicolumn{3}{c|}{Success rate (\%)} & \multicolumn{3}{c|}{LLM calls} & \multicolumn{3}{c|}{Tokens} & \multicolumn{3}{c}{Cost (\$)} \\
Benchmark & \hdtool & \hdcode & $\Delta$ & \hdtool & \hdcode & $\times$ & \hdtool & \hdcode & $\times$ & \hdtool & \hdcode & $\times$ \\
\midrule
LIBERO-PRO & 83.0 & 94.0 & \gain{+11.0} & 21.4 & 8.4 & \gain{2.5$\times$} & 1.23M & 273k & \gain{4.5$\times$} & 2.49 & 0.79 & \gain{3.1$\times$} \\
RoboTwin 2.0 & 60.0 & 69.2 & \gain{+9.2} & 12.2 & 8.9 & \gain{1.4$\times$} & 415k & 245k & \gain{1.7$\times$} & 1.00 & 0.66 & \gain{1.5$\times$} \\
RC365 atomic & 78.9 & 86.7 & \gain{+7.8} & 10.4 & 5.6 & \gain{1.8$\times$} & 261k & 171k & \gain{1.5$\times$} & 0.62 & 0.54 & \gain{1.2$\times$} \\
RC365 composite & 34.4 & 39.4 & \gain{+5.0} & 28.4 & 14.9 & \gain{1.9$\times$} & 1.22M & 603k & \gain{2.0$\times$} & 2.01 & 1.25 & \gain{1.6$\times$} \\
\midrule
All & 63.1 & 71.7 & \gain{\textbf{+8.6}} & 17.0 & 8.7 & \gain{\textbf{2.0$\times$}} & 788k & 276k & \gain{\textbf{2.9$\times$}} & 1.63 & 0.74 & \gain{\textbf{2.2$\times$}} \\
\bottomrule
\end{tabular}

  \end{adjustbox}
\end{table}

\begin{figure}[t]
  \centering
  \includegraphics[width=\linewidth]{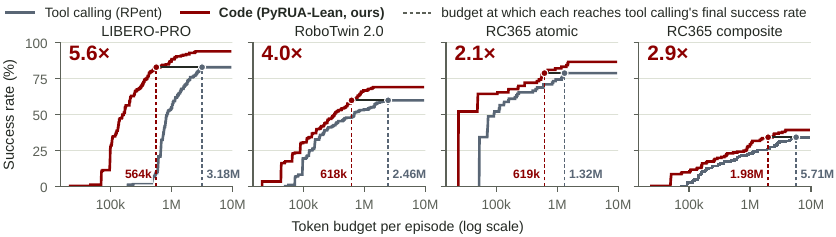}
  \caption{Success rate if every episode were stopped once its token consumption reached the budget on the horizontal
    axis. Dashed lines mark the budget at which each agent first reaches tool calling's final success rate, with its
    value on the axis; the large number is their ratio. RC365: RoboCasa365.}
  \label{fig:budget}
  
\end{figure}

\begin{figure}[t]
  \centering
  \includegraphics[width=\linewidth]{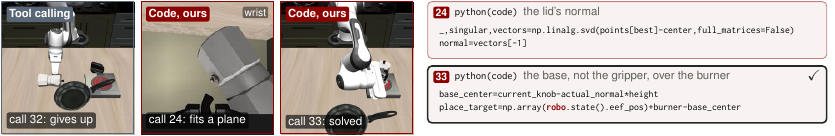}
  \caption{Only code solved this LIBERO-PRO instance (``turn on the stove and put the moka pot on it''): both agents
    dropped the pot, and the code agent fitted a plane to the lid, turned the pot upright and set its base over the
    burner (Appendix~\ref{app:pairs}).}
  \label{fig:moka}
\end{figure}

\subsection{Task Success and Inference Cost}
\label{sec:main_results}

\textbf{Task success improves across all benchmark groups.} Table~\ref{tab:main} and Figure~\ref{fig:overview}(a--c) show that PyRUA-Lean increases overall success from 63.1\% to 71.7\%, an improvement of 8.6 percentage points, or approximately 14\% relative. The gains are 11.0 points on LIBERO-PRO, 9.2 on RoboTwin 2.0, 7.8 on RoboCasa365 atomic tasks, and 5.0 on composite tasks. PyRUA-Lean alone solves 96 instances, whereas tool calling alone solves 36. These results support Hypothesis~1 within the evaluated setting: PyRUA-Lean completes more tasks under the same LLM-call budget. Figure~\ref{fig:moka} illustrates a recovery example in which PyRUA-Lean uses geometric computation to reposition a dropped object and complete the task. Appendix~\ref{app:pairs} provides further analysis of instances solved by only one agent.  

\textbf{Completing the same tasks requires fewer LLM calls and tokens.}
On jointly solved instances, PyRUA-Lean reduces mean LLM calls from 17.0 to 8.7 and input tokens from 788k to 276k, corresponding to reductions of 49\% and 65\%. Estimated cost decreases from \$1.63 to \$0.74 per episode. The smaller monetary reduction reflects, in part, caching discounts on the baseline's repeatedly processed history. These findings support Hypothesis~2 and show that the higher overall success rate is accompanied by lower inference overhead on shared successes. The pooled token reduction is influenced strongly by LIBERO-PRO's longer episodes; per-benchmark results are therefore also reported.

Figure~\ref{fig:budget} complements this comparison with retrospective token-budget cutoffs on recorded trajectories. On LIBERO-PRO, PyRUA-Lean reaches the baseline's final success rate with 564k tokens per episode, compared with 3.18M for tool calling. This connects success and efficiency beyond the jointly solved subset, although the agents were not rerun with these token budgets specified beforehand.

Section~\ref{sec:robodojo} evaluates the interface on a fourth benchmark, RoboDojo.

\Needspace{260pt}
\subsection{Analysis of LLM Calls and Observation Feedback}
\label{sec:efficiency_analysis}

\begin{wrapfigure}{r}{0.46\linewidth}
  \vspace{-22pt}
  \centering
  \includegraphics[width=\linewidth]{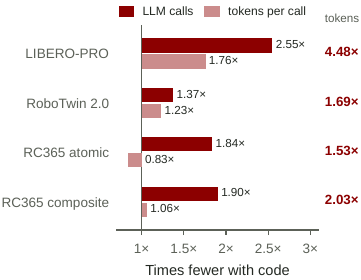}
  \caption{Token-savings decomposition on jointly solved instances. Tool-calling-to-PyRUA-Lean ratios quantify reductions in LLM calls (solid) and tokens per LLM call (light); their product gives the total token-reduction factor (right). Ratios below $1\times$ indicate increased usage by PyRUA-Lean.}
  \label{fig:split}
  \vspace{-8pt}
\end{wrapfigure}
\textbf{Fewer LLM calls account for most of the token reduction.}
Figure~\ref{fig:split} decomposes the token ratio into the LLM-call ratio and the ratio of average input tokens per LLM call. On LIBERO-PRO, the $4.48\times$ token ratio comprises $2.55\times$ fewer LLM calls and $1.76\times$ fewer tokens per LLM call. On RoboCasa365 atomic tasks, PyRUA-Lean's LLM calls are larger on average, yet fewer of them still reduce total token usage. Thus, Hypothesis~2 is supported primarily by reducing the number of LLM calls, rather than uniformly making each prompt smaller.

Execution traces illustrate how programmatic composition enables this reduction. A cell can execute dependent motions, check their outcomes, and retry a primitive before returning control to the LLM. In the bowl-placement example, the agent also computes placement geometry directly from observations. These behaviors address the central research question by moving intermediate coordination and computation into the execution runtime. The token decomposition is an \mbox{accounting} analysis, however, and does not independently isolate the causal contribution of each interface feature.

\textbf{On-demand images alone do not reproduce the gains.}
On LIBERO-PRO, we additionally modify the tool-calling baseline to return images only on request. As shown in Table~\ref{tab:observe}, success decreases from 83.0\% to 68.5\%. On the 131 instances solved by all three configurations, the modified baseline makes more LLM calls and consumes approximately the same total input tokens as the original baseline. This suggests that reducing automatic visual feedback alone is insufficient: observation requests must be considered together with action execution and replanning. The experiment does not separately quantify the contribution of selective feedback within PyRUA-Lean.

\subsection{Ablation Studies}
\label{sec:ablations}

\textbf{Token savings persist without VLA policies or guides.} We remove VLA policies, operating guides, or both from each agent on the same task instances (Appendix~\ref{app:ablation}); RoboCasa365 has no guides, so only VLA policies are ablated. Across these settings, PyRUA-Lean maintains higher overall success and lower input-token usage on jointly solved instances (Table~\ref{tab:ablation}), supporting Hypothesis~3. Without VLA policies, it achieves 68.8\% success on RoboTwin 2.0 versus 42.8\% for tool calling, with traces showing classical primitives composed into approach, gripper control, and state checks.

\textbf{Guidance has interface-dependent effects.} On RoboTwin 2.0 without VLA policies, removing guides increases tool-calling success from 42.8\% to 60.4\%, suggesting that guidance is not uniformly beneficial. Jointly solved subsets also vary across settings, so their token costs do not isolate component effects on a fixed task subset.

\begin{table}[t]
  \centering
  \caption{Ablations: the VLA policy, the operating guides or both removed from \emph{both} agents, on every task
    instance. Columns as in Table~\ref{tab:main}: success rate on all task instances, the other columns on the
    instances both agents solved in that setting; each group's first row is that table's. RoboCasa365 has no
    operating guides (--).}
  \label{tab:ablation}
  \footnotesize
  \setlength{\tabcolsep}{3.3pt}
  \begin{adjustbox}{max width=\linewidth}
  \begin{tabular}{@{}cc|ccc|ccc|ccc|ccc@{}}
\toprule
 & & \multicolumn{3}{c|}{Success rate (\%)} & \multicolumn{3}{c|}{LLM calls} & \multicolumn{3}{c|}{Tokens} & \multicolumn{3}{c}{Cost (\$)} \\
VLA & Guides & \hdtool & \hdcode & $\Delta$ & \hdtool & \hdcode & $\times$ & \hdtool & \hdcode & $\times$ & \hdtool & \hdcode & $\times$ \\
\midrule
\multicolumn{14}{c}{\textbf{\textit{LIBERO-PRO}}} \\
\abyes & \abyes & 83.0 & 94.0 & \gain{+11.0} & 21.4 & 8.4 & \gain{2.5$\times$} & 1.23M & 273k & \gain{4.5$\times$} & 2.49 & 0.79 & \gain{3.1$\times$} \\
\abyes & \abno & 84.0 & 88.0 & \gain{+4.0} & 19.0 & 9.3 & \gain{2.0$\times$} & 901k & 286k & \gain{3.1$\times$} & 1.89 & 0.71 & \gain{2.6$\times$} \\
\abno & \abyes & 73.0 & 80.5 & \gain{+7.5} & 22.9 & 10.9 & \gain{2.1$\times$} & 1.34M & 355k & \gain{3.8$\times$} & 2.60 & 0.86 & \gain{3.0$\times$} \\
\abno & \abno & 74.0 & 82.0 & \gain{+8.0} & 20.0 & 9.8 & \gain{2.0$\times$} & 943k & 290k & \gain{3.2$\times$} & 2.00 & 0.75 & \gain{2.7$\times$} \\
\midrule
\multicolumn{14}{c}{\textbf{\textit{RoboTwin 2.0}}} \\
\abyes & \abyes & 60.0 & 69.2 & \gain{+9.2} & 12.2 & 8.9 & \gain{1.4$\times$} & 415k & 245k & \gain{1.7$\times$} & 1.00 & 0.66 & \gain{1.5$\times$} \\
\abyes & \abno & 64.0 & 66.8 & \gain{+2.8} & 11.6 & 9.1 & \gain{1.3$\times$} & 392k & 251k & \gain{1.6$\times$} & 0.91 & 0.64 & \gain{1.4$\times$} \\
\abno & \abyes & 42.8 & 68.8 & \gain{+26.0} & 23.1 & 11.5 & \gain{2.0$\times$} & 1.03M & 316k & \gain{3.2$\times$} & 1.94 & 0.85 & \gain{2.3$\times$} \\
\abno & \abno & 60.4 & 66.4 & \gain{+6.0} & 16.9 & 12.0 & \gain{1.4$\times$} & 620k & 329k & \gain{1.9$\times$} & 1.29 & 0.88 & \gain{1.5$\times$} \\
\midrule
\multicolumn{14}{c}{\textbf{\textit{RoboCasa365 atomic}}} \\
\abyes & \abna & 78.9 & 86.7 & \gain{+7.8} & 10.4 & 5.6 & \gain{1.8$\times$} & 261k & 171k & \gain{1.5$\times$} & 0.62 & 0.54 & \gain{1.2$\times$} \\
\abno & \abna & 45.6 & 65.6 & \gain{+20.0} & 26.5 & 18.7 & \gain{1.4$\times$} & 760k & 627k & \gain{1.2$\times$} & 1.48 & 1.37 & \gain{1.1$\times$} \\
\midrule
\multicolumn{14}{c}{\textbf{\textit{RoboCasa365 composite}}} \\
\abyes & \abna & 34.4 & 39.4 & \gain{+5.0} & 28.4 & 14.9 & \gain{1.9$\times$} & 1.22M & 603k & \gain{2.0$\times$} & 2.01 & 1.25 & \gain{1.6$\times$} \\
\abno & \abna & 7.5 & 20.0 & \gain{+12.5} & 58.5 & 31.1 & \gain{1.9$\times$} & 2.58M & 1.14M & \gain{2.3$\times$} & 5.50 & 3.08 & \gain{1.8$\times$} \\
\bottomrule
\end{tabular}

  \end{adjustbox}
\end{table}

\subsection{Failure Cases and Limitations}
\label{sec:failure_analysis}

\textbf{Improved budget use explains some, but not all, additional successes.}
Of the 96 instances solved only by PyRUA-Lean, the baseline exhausts its LLM-call budget in 52 and ends unsuccessfully before exhausting it in 44. The former cases are consistent with programmatic composition enabling more execution within the budget. The latter include examples of geometric reasoning and scripted recovery, but also cases where a VLA succeeds in one run and fails in the other. The success difference therefore cannot be attributed entirely to better recovery logic.

\textbf{Efficiency gains depend on task structure and failures.}
Gains are smallest on RoboCasa365 atomic tasks, where a VLA policy often completes most of the task and the initial prompt accounts for much of the input. PyRUA-Lean's larger API description can offset part of the benefit from fewer LLM calls. Including failed episodes, tool calling uses \V{ROBOCASA_ALL_RATIO_1} the input tokens of PyRUA-Lean on RoboCasa365 atomic tasks, and the estimated inference cost of the two agents is about the same (Appendix~\ref{app:cost}). Efficiency on jointly solved instances should therefore be distinguished from the cost of all attempts.

\section{PyRUA-Lean on RoboDojo}
\label{sec:robodojo}

Section~\ref{sec:framework} holds the planner and the primitives fixed and varies the interface. This section asks a different question: how far does the interface carry a frozen LLM on a public benchmark built for pretrained robot policies? RoboDojo~\cite{chen2026robodojo} is such a benchmark, with a leaderboard led by VLA and world-action models~\cite{robodojo2026leaderboard}. Zhang et al.~\cite{zhang2026unexpected} recently showed that GPT-6 Astra, issuing one end-effector target per LLM call from RGB images, reaches \V{RD_ASTRA_SR}\% average success on it without any robot training, above every pretrained policy listed at the time but with almost no success where precision is needed. We evaluate the same LLM, on the same RGB observation, acting through PyRUA-Lean.

\subsection{Benchmark and Protocol}
\label{sec:rd_protocol}

RoboDojo's simulation track has 42 tabletop tasks for a dual-arm ARX-X5 station, grouped into five capability dimensions: Generalization (12 tasks), Precision (8), Long-Horizon (8), Memory (6), and Open (8). A task is evaluated on 50 public layouts per evaluation seed, and each seed has its own set of layouts; a Generalization task runs 25 episodes in its standard scene and 25 with randomized room and table appearance. One evaluation seed is therefore 2,100 episodes. The benchmark reports the success rate (SR), the percentage of episodes that meet the task's full-success criterion, and the Score, the mean process reward times 100, which gives partial credit for completed stages. As on the leaderboard, a task's values are averaged within its dimension, and the Average is the unweighted mean of the five dimension values.

We follow this protocol with RoboDojo's evaluation code (\texttt{scripts/robodojo.sh eval}, commit \texttt{cb2a37d}, with the XPolicyLab release it pins~\cite{xpolicylab2026}). Our agent is packaged as an XPolicyLab policy, like any policy the benchmark evaluates: RoboDojo's evaluation client runs the simulation, sends observations, executes the returned actions, and judges success, and it ends an episode as soon as the success check passes or the task's step limit is reached. An observation is three RGB images, from a head camera and from a camera on each wrist, and the joint and end-effector state of the two arms. We compare with the entries of the public leaderboard as of \V{RD_BOARD_DATE}. We evaluate on one evaluation seed (seed 0), 2,100 episodes; the configuration with depth of Section~\ref{sec:rd_depth} was evaluated on three.

\subsection{System}
\label{sec:rd_system}

One fixed agent runs all tasks. The planner is GPT-6 Astra at high reasoning effort, the model of Section~\ref{sec:framework} and of~\cite{zhang2026unexpected}; its weights are untouched and it sees no demonstrations. It acts only through \texttt{python(code)}, here in a minimal agent loop of our own around an OpenAI-compatible endpoint instead of the Codex CLI, with a budget of 100 LLM calls per episode (counted on the calls that run a code cell), images on demand, and no memory across episodes (settings in Appendix~\ref{app:robodojo}).

\begin{table}[t]
  \centering
  \caption{Our primitive library for RoboDojo: everything the agent's code can call. There is no VLA policy and no
    pick, place, or grasp primitive; SAM~3 is the only learned component besides the LLM. Every method works from the
    RGB images and the robot state that RoboDojo's evaluation client sends; none reads a depth image or the
    simulator's state.}
  \label{tab:rdprims}
  \footnotesize
  \setlength{\tabcolsep}{5pt}
  \begin{tabular}{@{}lp{0.3\linewidth}p{0.5\linewidth}@{}}
    \toprule
    & Method of \texttt{robo} & What it does \\
    \midrule
    Motion & \texttt{move\_to(xyz, arm, gripper, pitch, yaw, roll, path)} & Drives one arm's grasp point to a pose.
      The library solves the inverse kinematics; the path is a straight line of the grasp point or a joint-space
      swing. Reports whether the pose was reached, or which nearby poses are reachable. \\
    & \texttt{set\_gripper(arm, command, opening)} & Runs a full gripper stroke; reports whether the fingers stopped
      on something and how wide it is. \\
    & \texttt{home(arm)} & Returns an arm to its initial joint angles. \\
    \midrule
    Perception & \texttt{locate(text, camera)} & Segments the object named in text with SAM~3 in one camera's image,
      finds it in the other cameras, and measures it in 3D from these views: center, top, footprint, and the heading
      for a grasp across its narrow side. Reports how many cameras agreed: one camera cannot measure a height. \\
    & \texttt{segment(text, camera)}, \texttt{back\_project(row, col, camera)} & The mask of a named object; the world
      point at a pixel, on the object under it or on the table. \\
    \midrule
    Observation & \texttt{state()}, \texttt{show(camera)}, \texttt{image(camera)}, \texttt{artifact(name)},
      \texttt{camera\_meta(camera)} &
      End-effector poses, gripper openings and joint angles; attach a camera image to the cell's feedback; the image
      as an array or as file bytes; a camera's calibration (a wrist camera's follows from its arm's state). \\
    \bottomrule
  \end{tabular}
\end{table}

\textbf{Primitives.} Where Section~\ref{sec:framework} runs RPent's primitive implementations, the primitives of the RoboDojo agent are our own: we wrote the library for the benchmark's dual-arm station, and Table~\ref{tab:rdprims} lists everything in it that the code can call. There is no pick, place, or grasp primitive: a grasp is a few lines of code that locate an object, move above it with the measured heading, descend along a line, and close the gripper while checking the reported width. Without a depth image, \texttt{locate} measures an object from its masks in the cameras that see it: the fixed head camera bounds its footprint on the table, and a wrist camera that also sees it settles its height, so a program often moves a wrist camera above an object and measures again before it grasps. We added a primitive only when a program could not build its function from the others.

\textbf{No VLA policy.} Unlike the agents of Section~\ref{sec:framework}, this one has no VLA policy among its primitives, and that is a choice. A VLA policy is a stochastic primitive: the same call may succeed in one episode and fail in the next, and in Section~\ref{sec:framework} some of the instances that only one agent solved differ only in whether the same VLA call succeeded (Appendix~\ref{app:pairs}). We did not want this variance in the result. (The configuration with depth varies by a fraction of a point between evaluation seeds, Section~\ref{sec:rd_depth}.) The primitives of Table~\ref{tab:rdprims} are geometric and report what they measured or reached, so a program can check every step, and what the result measures is the programs the LLM writes. As in the LLM-as-policy setting of~\cite{zhang2026unexpected}, no learned policy acts on the robot; unlike that setting, our agent has a perception primitive built on a segmentation model~\cite{carion2025sam3}.

\textbf{No privileged information.} During an episode, everything the agent observes is the observation that RoboDojo's evaluation client sends to every policy: the three RGB images and the joint and end-effector state. It reads no depth image and nothing of the simulator's scene state: it is given no object poses and no description of the layout, so it measures where things are from the images, with \texttt{locate}. What the library knows in advance is the station itself, that is, the calibration of its cameras, where its arms stand, and the height of its table, which are the same in every episode. It also receives no success or progress signal while an episode runs: unlike the agents of Section~\ref{sec:framework}, which can read the benchmark's success flag, it has to judge from the images and the robot state whether the task is done.

\textbf{Prompt.} The prompt holds the API reference, operating notes on how the tools behave (frames, reach, how the gripper reports a grasp, how to measure an object with the wrist cameras) and on how strictly the benchmark judges a final scene, and a task card with the task's name, the instruction the benchmark generates for the layout, and the first camera images. Beyond that instruction, it contains no description of a task and no recipe for solving one. The direct-control agent of~\cite{zhang2026unexpected}, in contrast, receives for each task a description and its scoring stages, derived from the benchmark's wiki.

\subsection{Results}
\label{sec:rd_results}

\begin{table}[t]
  \centering
  \caption{RoboDojo-Sim in the leaderboard's format; each value is Score\,/\,SR (\%). \emph{First row:} PyRUA-Lean
    (ours) over \V{RD_SEED_PHRASE} (\V{RD_EPISODES} episodes). \emph{Other rows:} \V{RD_BOARD_SHOWN_PHRASE}
    of the public leaderboard~\cite{robodojo2026leaderboard} as of \V{RD_BOARD_DATE}, in its order (by Average Score,
    \V{RD_BOARD_N} entries). Gen.\ is the mean of the standard and randomized halves; WAM: world-action model. Bold: the
    highest of a column.}
  \label{tab:rdboard}
  \footnotesize
  \setlength{\tabcolsep}{3.5pt}
  \begin{adjustbox}{max width=\linewidth}
  \begin{tabular}{@{}clc|c|ccccc@{}}
\toprule
Rank & Entry & Type & Average & Gen. & Prec. & Long-Hor. & Memory & Open \\
\midrule
\rowcolor{gaintint} -- & \textbf{PyRUA-Lean} (ours, GPT-6 Astra) & Agent & \textbf{58.20\,/\,51.62} & \textbf{50.65\,/\,45.83} & \textbf{47.02\,/\,37.25} & \textbf{63.26\,/\,48.00} & \textbf{63.00\,/\,63.00} & \textbf{67.09\,/\,64.00} \\
\midrule
1 & VPP2~\cite{guo2026vpp2} & WAM & 39.26\,/\,32.26 & 36.38\,/\,29.50 & 43.03\,/\,35.17 & 56.28\,/\,37.00 & 56.01\,/\,55.56 & 4.61\,/\,4.08 \\
2 & PhysicalRSI~\cite{physicalrsi2026} & Agent + VLA & 36.27\,/\,31.38 & 21.30\,/\,15.62 & 38.05\,/\,32.50 & 46.37\,/\,37.33 & 46.74\,/\,46.56 & 28.88\,/\,24.92 \\
3 & Awomo-0.5~\cite{awomo2026awomo05} & WAM & 35.34\,/\,29.64 & 30.31\,/\,23.84 & 39.65\,/\,33.50 & 40.36\,/\,26.75 & 42.27\,/\,41.22 & 24.09\,/\,22.92 \\
4 & Simate-beta~\cite{simate2026beta} & VLA & 33.95\,/\,27.96 & 35.09\,/\,27.95 & 34.35\,/\,26.92 & 57.84\,/\,43.42 & 33.33\,/\,33.00 & 9.12\,/\,8.50 \\
5 & Rex-M1 Preview~\cite{rexm12026preview} & VLA & 33.79\,/\,27.44 & 34.05\,/\,26.05 & 32.82\,/\,23.83 & 46.89\,/\,33.00 & 48.53\,/\,48.22 & 6.67\,/\,6.08 \\
6 & VPP2-Preview~\cite{guo2026vpp2} & WAM & 31.40\,/\,25.62 & 31.34\,/\,24.62 & 33.75\,/\,25.58 & 35.31\,/\,22.17 & 50.78\,/\,50.33 & 5.83\,/\,5.42 \\
7 & InternW0-$\Delta$~\cite{internrobotics2026internw0delta} & WAM & 30.77\,/\,23.91 & 30.09\,/\,22.78 & 31.98\,/\,23.25 & 46.29\,/\,29.33 & 34.67\,/\,34.00 & 10.84\,/\,10.17 \\
8 & Liber-0 Preview~\cite{liberai2026liber0} & VLM + WAM & 30.74\,/\,25.52 & 24.99\,/\,18.61 & 38.28\,/\,33.17 & 45.98\,/\,33.17 & 37.77\,/\,37.33 & 6.68\,/\,5.33 \\
9 & Liber-0 Lite~\cite{liberai2026liber0} & WAM & 29.24\,/\,24.23 & 25.36\,/\,19.11 & 36.60\,/\,31.08 & 45.50\,/\,32.92 & 35.68\,/\,35.44 & 3.06\,/\,2.58 \\
10 & GPT-6 Astra, direct control~\cite{zhang2026unexpected} & Agent & 28.97\,/\,22.48 & 33.36\,/\,30.50 & 12.65\,/\,4.00 & 21.45\,/\,8.25 & 43.04\,/\,38.67 & 34.36\,/\,31.00 \\
\bottomrule
\end{tabular}

  \end{adjustbox}
\end{table}

\begin{figure}[t]
  \centering
  \includegraphics[width=\linewidth]{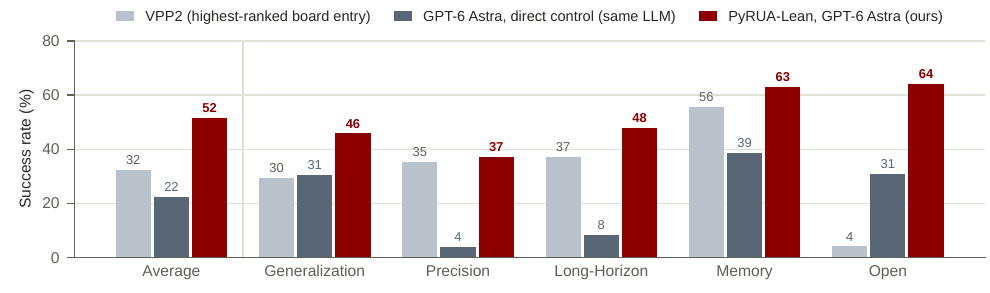}
  \caption{Success rate on RoboDojo, on average and per capability dimension: the highest-ranked entry of the public
    board, GPT-6 Astra controlling the end-effector directly~\cite{zhang2026unexpected}, and the same LLM acting
    through PyRUA-Lean (ours, \V{RD_SEED_PHRASE}).\V{RD_STAR_NOTE}}
  \label{fig:rddims}
\end{figure}

\textbf{PyRUA-Lean is above every entry of the board.} Over \V{RD_EPISODES} episodes (\V{RD_SEED_PHRASE}), the agent reaches an Average SR of \V{RD_SR}\% and a Score of \V{RD_SCORE} (Table~\ref{tab:rdboard}); pooled over all episodes, it solves \V{RD_MICRO_SOLVED} of \V{RD_MICRO_N} (\V{RD_MICRO_SR}\%). The highest-ranked entry of the board, \V{RD_TOP_NAME}, has \V{RD_TOP_SR}\% and \V{RD_TOP_SCORE}; PhysicalRSI, the strongest agent entry, has \V{RD_RSI_SR}\% and \V{RD_RSI_SCORE}. PyRUA-Lean has the highest success rate on \V{RD_DIMS_BEST_PHRASE} dimensions (Table~\ref{tab:rdboard}), and on \V{RD_TASK_WINS} of the \V{RD_TASKS_N} tasks it is above every entry of the board (Appendix~\ref{app:rd_tasks}).

\textbf{Same LLM, another interface.} Against GPT-6 Astra controlling the end-effector directly, the Average SR rises from \V{RD_ASTRA_SR}\% to \V{RD_SR}\%, by a factor of \V{RD_VS_ASTRA_X} (Figure~\ref{fig:rddims}). The largest differences are where direct control is weakest: Precision goes from \V{RD_ASTRA_PREC_SR}\% to \V{RD_PREC_SR}\% and Long-Horizon from \V{RD_ASTRA_LONG_SR}\% to \V{RD_LONG_SR}\%. PyRUA-Lean is higher on \V{RD_ABOVE_ASTRA_N} of the \V{RD_TASKS_N} tasks and lower on \V{RD_BELOW_ASTRA_N}. The two systems share the LLM and the RGB observation and differ in several things at once: the action interface (programs over primitives against one end-effector target per LLM call), perception (a segmentation-based \texttt{locate} against the images as the LLM reads them), reasoning effort (high against medium), and the prompt (tool notes against per-task descriptions). The gap therefore measures the systems as a whole, not the observation, which is the same.

\begin{table}[t]
  \centering
  \caption{Tasks with the largest differences between PyRUA-Lean and the best entry of the public board on that task
    (SR in \%; the board gives per-task results for 57 policies and ranks \V{RD_BOARD_N}). Appendix~\ref{app:rd_tasks} has all \V{RD_TASKS_N} tasks.}
  \label{tab:rdmargins}
  \footnotesize
  \begin{tabular}{@{}lrrl@{}}
\toprule
Task & Ours & Best & Best board entry \\
\midrule
\multicolumn{4}{@{}l}{\emph{Largest margins over the best board entry}} \\
\texttt{imitate\_sorting\_sequence} & 98.0 & 36.0 & GPT-6 Astra (direct) \\
\texttt{classify\_objects\_by\_language} & 92.0 & 30.0 & GPT-6 Astra (direct) \\
\texttt{stack\_blocks\_by\_language} & 100.0 & 40.0 & GPT-6 Astra (direct) \\
\texttt{sort\_nesting\_dolls\_by\_size} & 74.0 & 18.7 & VPP2 \\
\texttt{play\_stacking\_toy} & 66.0 & 23.3 & VPP2 \\
\texttt{classify\_objects} & 98.0 & 56.0 & GPT-6 Astra (direct) \\
\midrule
\multicolumn{4}{@{}l}{\emph{Largest deficits to the best board entry}} \\
\texttt{press\_by\_number} & 0.0 & 96.0 & VPP2 \\
\texttt{make\_kong} & 0.0 & 90.0 & GalaxeaVLA (G0.5) \\
\texttt{build\_tower} & 2.0 & 83.3 & Simate-beta \\
\texttt{pour\_liquid\_into\_cup} & 8.0 & 67.3 & Rex-M1 Preview \\
\texttt{insert\_key} & 0.0 & 39.3 & PhysicalRSI \\
\texttt{swap\_blocks} & 0.0 & 38.7 & Simate-beta \\
\bottomrule
\end{tabular}

\end{table}

\textbf{Where it fails.} \V{RD_FULL_N} tasks are solved in every episode (\V{RD_FULL_TASKS}), and \V{RD_ZERO_N} in none (\V{RD_ZERO_TASKS}). The failures are not spread evenly (Table~\ref{tab:rdmargins}): pretrained policies reach 83--96\% on pressing buttons a given number of times (\texttt{press\_by\_number}), on a tile game against an opponent (\texttt{make\_kong}), and on building a tower from blocks and boards (\texttt{build\_tower}), where our agent is at or near zero. The other tasks without a success ask for a key to be handed over, inserted and turned, for a button to be pressed after each move, or for objects to be stored in stands and fitted compartments. Several of these tasks involve contact whose outcome the agent's feedback does not report, such as a button press, or fits that are tight relative to the accuracy of its perception. We have not analyzed these failures episode by episode.

\textbf{Cost.} An episode takes \V{RD_LLM_MEAN} LLM calls on average, \V{RD_LLM_SOLVED} when it is solved and \V{RD_LLM_FAILED} when it is not; the longest took \V{RD_LLM_MAX}, and no episode ran more than \V{RD_CELLS_MAX} code cells, so none reached the budget of 100. An episode consumes \V{RD_INTOK_MEAN_K}k input tokens, cached input included, and \V{RD_OUTTOK_MEAN_K}k output tokens, and lasts about \V{RD_WALL_MEAN_MIN} minutes. At about \$\V{RD_USD_EPISODE} per episode at our model gateway, the 2,100 episodes of the seed cost about \$\V{RD_USD_SEED}. At this cost we evaluated two configurations, this one and the one with depth of Section~\ref{sec:rd_depth}, which ran on three seeds; we did not run the evaluation with a VLA policy added to the primitives.

\subsection{The Same Agent with Depth}
\label{sec:rd_depth}

\begin{table}[t]
  \centering
  \caption{PyRUA-Lean on RoboDojo from RGB images alone, and with the simulator's depth added to the observation:
    each value Score\,/\,SR (\%), as in Table~\ref{tab:rdboard}. The configuration with depth was evaluated on three
    seeds; the standard deviation over them is the population one, as in RoboDojo's summary. The configuration with depth is not compared with the leaderboard.}
  \label{tab:rddepth}
  \footnotesize
  \setlength{\tabcolsep}{4pt}
  \begin{adjustbox}{max width=\linewidth}
  \begin{tabular}{@{}lr|c|ccccc@{}}
\toprule
Observation & Episodes & Average & Gen. & Prec. & Long-Hor. & Memory & Open \\
\midrule
\multicolumn{8}{@{}l}{\emph{The same evaluation seed}} \\
RGB (Section~\ref{sec:rd_results}) & 2{,}100 & 58.20\,/\,51.62 & 50.65\,/\,45.83 & 47.02\,/\,37.25 & 63.26\,/\,48.00 & 63.00\,/\,63.00 & 67.09\,/\,64.00 \\
RGB and depth & 2{,}100 & 58.59\,/\,52.42 & 49.71\,/\,43.83 & 49.92\,/\,41.00 & 63.02\,/\,49.00 & 64.00\,/\,64.00 & 66.29\,/\,64.25 \\
\midrule
\multicolumn{8}{@{}l}{\emph{Three evaluation seeds}} \\
RGB and depth, mean & 6{,}300 & 58.81\,/\,52.91 & 51.11\,/\,45.44 & 49.78\,/\,41.08 & 64.73\,/\,51.75 & 64.02\,/\,64.00 & 64.39\,/\,62.25 \\
\quad standard deviation over the seeds & & 0.15\,/\,0.35 & 1.10\,/\,1.23 & 0.57\,/\,0.51 & 1.39\,/\,2.15 & 0.02\,/\,0.00 & 2.16\,/\,2.16 \\
\bottomrule
\end{tabular}

  \end{adjustbox}
\end{table}

RoboDojo's evaluation client can also add a depth image and the matrices of each camera to the observation. The leaderboard does not admit this depth for agent entries, so the results above do not use it and this configuration is not compared with the board. With depth, \texttt{locate} measures an object from the depth pixels under its mask, \texttt{back\_project} reads the depth at a pixel, and \texttt{camera\_meta} returns the matrices the client sends; the notes on measuring with the wrist cameras are dropped, and nothing else changes (Appendix~\ref{app:rd_depth}).

Depth helps: on \V{RDX_SCOPE}, it raises the Average SR from \V{RDX_RGB_SR}\% to \V{RDX_D_SR}\% and the Score from \V{RDX_RGB_SCORE} to \V{RDX_D_SCORE}, most on \V{RDX_MAXDIFF_DIM} (\V{RDX_MAXDIFF_RGB}\% to \V{RDX_MAXDIFF_D}\%; Table~\ref{tab:rddepth}). Over \V{RDD_SEED_PHRASE} (\V{RDD_EPISODES} episodes), the configuration with depth reaches an Average SR of \V{RDD_SR}\% and a Score of \V{RDD_SCORE}. The seeds, which have different layouts, give \V{RDD_SEED_SRS}, a standard deviation of \V{RDD_SR_SD} points: one seed is a close estimate of the success rate, and the lead over the leaderboard in Section~\ref{sec:rd_results} is not an effect of the seed.

\section{PyRUA-Lean Coder: Training an Open Model to Use the Interface}
\label{sec:coder}

Sections~\ref{sec:framework} and~\ref{sec:robodojo} use a frontier LLM behind an API. This section asks whether a small open-weight model can learn to act through the same interface. We train the planner itself: the LLM that reads the task, writes a cell, reads its output and the images it asked for, and decides what to do next. The primitives, including the frozen VLA policies, stay as in Section~\ref{sec:framework}. We call the trained planner \textbf{PyRUA-Lean Coder v0.1-4B}.

\subsection{Training}
\label{sec:coder_training}

\textbf{Model.} We start from Qwen3.5-4B~\cite{qwen2026qwen35}, an open-weight vision-language model with 4B parameters that reasons before it answers.

\textbf{Episodes and reward.} A training episode is the interaction of Section~\ref{sec:framework}: the prompt with the API reference, the operating guides and the task card (instruction, initial robot state, first camera images), followed by alternating LLM calls (the model reasons, then calls \texttt{python}) and tool results (text, and the images the code requested). The reward is 1 if the benchmark's success check passes and 0 otherwise, with no shaping and no penalty for LLM calls or simulator steps. To keep rollouts short, a training episode is limited to 24 code cells and 25 minutes, against 40 cells and 90 minutes when the same benchmark is evaluated, and an LLM call to 4,096 new tokens.

\textbf{Training instances.} We train on 45 LIBERO-PRO task instances: three tasks from each of the goal, spatial, and object suites at five seeds. The other 31 LIBERO-PRO tasks, among them the whole long-horizon suite, and all of RoboTwin 2.0 and RoboCasa365 are never used.

\textbf{Algorithm.} We use group relative policy optimization (GRPO)~\cite{shao2024deepseekmath}, a reinforcement-learning algorithm. In each iteration, 16 instances are drawn and the current model plays each of them 8 times, 128 episodes in all. An episode's advantage is its reward normalized within its group of 8, and the clipped policy-gradient loss (clip range 0.2 below and 0.28 above, as in~\cite{yu2025dapo}) is applied to every token the model generated in the episode, reasoning and code alike. An iteration makes two optimizer steps of 64 episodes each, with Adam at a constant learning rate of $10^{-6}$ and without a KL or an entropy term. The model is fine-tuned in full, without adapters, starting from the base weights: there is no supervised stage and no demonstration data. A group that loses an episode to an infrastructure failure is dropped and sampled again. Training runs for 30 iterations, that is 3,840 episodes and 60 optimizer steps, and the checkpoint after the last iteration is the one we evaluate and will release (Appendix~\ref{app:coder} lists all settings).

\textbf{Infrastructure.} Training uses the miles framework~\cite{radixark2026miles} with SGLang~\cite{zheng2023sglang} serving the rollouts, on one node with eight GPUs; the simulators and the frozen VLA and SAM~3 services run next to it. The run takes about twelve hours.

\subsection{Evaluation Protocol}
\label{sec:coder_protocol}

We evaluate on the 700 task instances of Section~\ref{sec:framework}, one episode per instance, with that section's prompt, tools, primitives, frozen VLA policies and budgets. A budget counts the code cells of an episode, 40 or, on RoboCasa365 composite tasks, 100; GPT-6 Astra writes one cell per LLM call, which makes it the LLM-call budget of Section~\ref{sec:framework}. The runs differ from Section~\ref{sec:framework}'s GPT-6 Astra runs in the agent loop, which is PyRUA-Lean's own instead of the Codex CLI and requires a tool call in every LLM call; in the context, which is 65,536 tokens (an episode that exceeds it ends there); and in the wall-clock limit, which is 90 minutes. However an episode ends, the benchmark's success check decides whether it is solved. Sampling follows the model card's settings for reasoning (temperature 1.0, top-$p$ 0.95, top-$k$ 20), with at most 8,192 new tokens per LLM call. The base model, the trained model and Qwen3.5-27B~\cite{qwen2026qwen35b27} are evaluated identically.

LIBERO-PRO's 200 instances contain the 45 training instances, so we report them separately from the 155 instances of tasks the model was not trained on. Together with RoboTwin 2.0 and RoboCasa365, these give 655 instances that training never saw. An instance whose episode failed on infrastructure is run again and does not count until it has a result.\V{CD_FORCED_NOTE}

\subsection{Results}
\label{sec:coder_results}

\begin{table}[t]
  \centering
  \caption{Success rate (\%) of Qwen3.5-4B before and after training (PyRUA-Lean Coder v0.1-4B), and of
    Qwen3.5-27B, on the 700 task instances of Section~\ref{sec:framework}, one episode per instance. Training used 45 LIBERO-PRO
    instances only.\V{CD_COUNT_NOTE}
    \emph{Last row:} GPT-6 Astra with PyRUA-Lean from Table~\ref{tab:main}, with its LIBERO-PRO result split in the
    same way, run in a different agent loop (Section~\ref{sec:coder_protocol}), for reference. Bold: the highest value of
    a column among the first three rows. RC365: RoboCasa365.}
  \label{tab:coder}
  \footnotesize
  \setlength{\tabcolsep}{5pt}
  \begin{adjustbox}{max width=\linewidth}
  \begin{tabular}{@{}l|c|cccc|c|c@{}}
\toprule
 & Trained on & \multicolumn{5}{c|}{Not seen in training} & \\
Planner & \makecell{LIBERO-PRO\\45 instances} & \makecell{LIBERO-PRO\\other tasks (155)} & \makecell{RoboTwin 2.0\\(250)} & \makecell{RC365\\atomic (90)} & \makecell{RC365\\composite (160)} & \makecell{All unseen\\(655)} & \makecell{All\\(700)} \\
\midrule
Qwen3.5-4B & 37.8 & 13.5 & 14.0 & 52.2 & 12.5 & 18.8 & 20.0 \\
Qwen3.5-27B & 64.4 & \textbf{38.1} & 20.8 & 65.6 & 13.1 & 29.2 & 31.4 \\
\rowcolor{gaintint} \textbf{PyRUA-Lean Coder v0.1-4B} & \textbf{71.1} & 37.4 & \textbf{52.4} & \textbf{76.7} & \textbf{19.4} & \textbf{44.1} & \textbf{45.9} \\
\midrule
\textcolor{black!55}{GPT-6 Astra (Section~\ref{sec:framework})} & \textcolor{black!55}{100.0} & \textcolor{black!55}{92.3} & \textcolor{black!55}{69.2} & \textcolor{black!55}{86.7} & \textcolor{black!55}{39.4} & \textcolor{black!55}{69.8} & \textcolor{black!55}{71.7} \\
\bottomrule
\end{tabular}

  \end{adjustbox}
\end{table}

\textbf{GRPO on 45 instances lifts the 4B model on every benchmark.} On the training instances, Qwen3.5-4B goes from \V{CD_B4_TRAIN}\% to \V{CD_C4_TRAIN}\% (Table~\ref{tab:coder}). On LIBERO-PRO tasks it was not trained on, it goes from \V{CD_B4_HELD}\% to \V{CD_C4_HELD}\%; on RoboTwin 2.0 from \V{CD_B4_RT}\% to \V{CD_C4_RT}\%; on RoboCasa365 atomic tasks from \V{CD_B4_RCA}\% to \V{CD_C4_RCA}\%; and on composite tasks from \V{CD_B4_RCC}\% to \V{CD_C4_RCC}\%. Over all instances not seen in training, the success rate rises from \V{CD_B4_UNSEEN}\% to \V{CD_C4_UNSEEN}\% (Figure~\ref{fig:overview}(e)). The trained 4B model is thereby above Qwen3.5-27B, which reaches \V{CD_B27_UNSEEN}\% on these instances under the same protocol.

\textbf{The gain transfers across simulators, robots, and VLA policies.} RoboTwin 2.0 is a dual-arm benchmark in another simulator, with LingBot-VLA as its policy primitive; RoboCasa365 has a mobile manipulator in kitchens, with RLDX-1. Neither appears in training, which saw one single-arm robot and $\pi_{0.5}$ on nine tabletop tasks. The transfer is therefore not explained by seeing the evaluation tasks, scenes, or robots.

\begin{table}[t]
  \centering
  \caption{How episodes run and end, on the instances not seen in training (means, except code cells per LLM call,
    a median). \emph{Code cells} counts the cells a model wrote, including those refused once its budget was spent.
    \emph{Images requested} leaves out the two or three pictures of the task card, which every episode starts with.
    \emph{Gave up}: the model ended the episode itself without success; \emph{Cell budget}, \emph{Wall clock},
    \emph{Context overflow}: the episode was stopped at that limit without success. Every episode is in exactly one
    of the five columns on the right.\V{CD_BEH_LIMIT_NOTE}\V{CD_BEH_NOTE}}
  \label{tab:coderbeh}
  \footnotesize
  \setlength{\tabcolsep}{4.5pt}
  \begin{tabular}{@{}lrrrr|rrrrr@{}}
\toprule
 & \multicolumn{4}{c|}{Per episode} & \multicolumn{5}{c}{How the episode ended (\%)} \\
Planner & \makecell{LLM\\calls} & \makecell{Code cells per\\LLM call} & \makecell{Code\\cells} & \makecell{Images\\requested} & Solved & \makecell{Gave\\up} & \makecell{Cell\\budget} & \makecell{Wall\\clock} & \makecell{Context\\overflow} \\
\midrule
Qwen3.5-4B & 9.4 & 9.0 & 89.4 & 19.3 & 18.8 & 6.9 & 52.7 & 2.4 & 19.2 \\
Qwen3.5-27B & 35.5 & 1.0 & 34.9 & 22.5 & 29.2 & 4.6 & 35.0 & 0.3 & 31.0 \\
\textbf{PyRUA-Lean Coder v0.1-4B} & 21.1 & 1.0 & 21.0 & 0.8 & 44.1 & 25.2 & 10.5 & 16.8 & 3.4 \\
\bottomrule
\end{tabular}

\end{table}

\textbf{What changes is how the model uses the interface.} Table~\ref{tab:coderbeh} shows the three models on the unseen instances. The base model writes \V{CD_B4_BEH_PER_CALL} code cells per LLM call (median over its episodes; more than 1.5 in \V{CD_B4_BEH_MULTI}\% of them): it commits a batch of actions before seeing the result of the first. Its episodes last \V{CD_B4_BEH_LLMCALLS} LLM calls on average; \V{CD_B4_BEH_BUDGET}\% of them end on the budget of code cells and \V{CD_B4_BEH_OVERFLOW}\% on a context overflow. Qwen3.5-27B writes one cell per LLM call, but it requests \V{CD_B27_BEH_IMG} images per episode and ends \V{CD_B27_BEH_BUDGET}\% of its episodes on the cell budget and \V{CD_B27_BEH_OVERFLOW}\% on a context overflow. After training, the 4B model writes one cell per LLM call and reads its result before the next. It requests \V{CD_C4_BEH_IMG} images per episode instead of \V{CD_B4_BEH_IMG}, and endings on the cell budget and on a context overflow fall to \V{CD_C4_BEH_BUDGET}\% and \V{CD_C4_BEH_OVERFLOW}\%. In exchange, more episodes end because the model gives up (\V{CD_C4_BEH_GAVEUP}\%) or runs into the wall clock (\V{CD_C4_BEH_TIMEOUT}\%). This is the feedback-driven use of the interface described in Section~\ref{sec:primitive_composition}, which the frontier model shows without training; it would also explain why a gain learned on nine tasks carries over to other benchmarks. We do not claim that training improved manipulation skill as such: grasps and motions are still executed by the same primitives and frozen VLA policies.

\begin{figure}[t]
  \centering
  \includegraphics[width=0.42\linewidth]{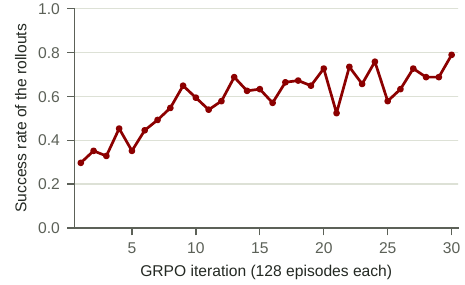}
  \caption{Training of PyRUA-Lean Coder v0.1-4B: success rate of the rollouts in each GRPO iteration (16 training
    instances $\times$ 8 episodes, under the training limit of 24 code cells).}
  \label{fig:codertraining}
\end{figure}

\textbf{Training dynamics.} The success rate of the training rollouts rises from \V{CD_TR_FIRST} in the first iteration to \V{CD_TR_LAST5} over the last five (Figure~\ref{fig:codertraining}). These rollouts run on the 45 training instances and under the shorter training limits, so they are not comparable with the evaluation numbers above.

\textbf{Distance to the frontier model.} PyRUA-Lean Coder v0.1-4B solves \V{CD_C4_ALL}\% of the 700 instances, against \V{CD_GPT_ALL}\% for GPT-6 Astra with the same interface and primitives in Section~\ref{sec:framework}. The two numbers come from different agent loops and limits and are not a controlled comparison, but the gap is large. It is smallest on RoboCasa365 atomic tasks and RoboTwin 2.0 and largest on unseen LIBERO-PRO tasks (Table~\ref{tab:coder}).

\section{Limitations and Future Work}
\label{sec:limitations}

\textbf{The interface comparison.} The comparison of Section~\ref{sec:framework} is limited to GPT-6 Astra, RPent's robot stacks and primitive libraries, and simulation, with one run per agent per task instance. Generalization to other planners and primitive configurations, variability across repeated runs, and physical-robot performance remain untested. The comparison evaluates the complete interfaces without fully isolating the effects of primitive composition, persistent state, and selective feedback.

\textbf{RoboDojo.} The RoboDojo agent is a system around a frontier LLM behind an API, with a segmentation model, where most leaderboard entries are single pretrained policies, and we have not measured how much of its margin over direct control comes from the perception primitive and how much from programs. Its failures concentrate on tasks that involve contact it gets no feedback on, or tight fits, and we have not analyzed them episode by episode. We left VLA policies out of the agent because of their run-to-run variance (Section~\ref{sec:rd_system}). An episode costs dollars and minutes, which is why we evaluated only two configurations, with and without depth, and have not measured what a VLA policy would add as a further primitive.

\textbf{PyRUA-Lean Coder.} PyRUA-Lean Coder v0.1-4B is a first version: one model size and one training run, on 45 instances of one benchmark and with success as the only reward. What the model demonstrably learned is to use the interface step by step, and it remains far below the frontier model on the same tasks. We evaluated it with the primitives and VLA policies of Section~\ref{sec:framework} only; whether it can drive the VLA-free primitives of Section~\ref{sec:robodojo} is untested.

Future work will address these limitations, measure execution latency and recovery on physical robots, examine how primitive granularity and cross-episode reuse of validated routines affect success, inference cost, and adaptability, and train larger models, on more benchmarks and for longer.


\section{Conclusion}
\label{sec:conclusion}

We presented PyRUA-Lean, an interactive code-execution interface that composes existing robot primitives, including VLA policies, and controls which execution feedback enters the LLM context. Across 700 simulated task instances, PyRUA-Lean outperforms a tool-calling baseline using the same GPT-6 Astra planner and underlying primitives under equal LLM-call budgets, increasing overall success from 63.1\% to 71.7\%, and on instances solved by both agents it uses 49\% fewer LLM calls and 65\% fewer input tokens. With a handful of geometric primitives, no learned policy acting on the robot, and no privileged information, the same planner reaches \V{RD_SR}\% average success rate on RoboDojo, above every entry of the public leaderboard. And the interface can be learned: GRPO on 45 task instances turns an open 4B model into a planner that solves \V{CD_C4_UNSEEN}\% instead of \V{CD_B4_UNSEEN}\% of the instances it never trained on, across three benchmarks. These results show the value of programmatic primitive composition for frontier and small models alike, and motivate further study of feedback management and of training in robot agents.

\clearpage

\bibliographystyle{plainnat}
\setlength{\bibhang}{0pt}
\setlength\bibindent{0pt}
\bibliography{references}


\clearpage
\appendix

{
  \centering
  \textbf{\Large Fewer Tokens, Better Action: GPT-6 Astra Robot Agents with \V{POOLED_SR_REL} Higher Success Rate
  but \V{POOLED_TOKENS_FEWER} Fewer Tokens}

  \vspace{5pt}
  {\Large Appendix}
  \par
}

\section{The episode of Figure~\ref{fig:method}, LLM call by LLM call}
\label{app:calls}

\begin{figure}[h]
  \centering
  \includegraphics[width=\linewidth]{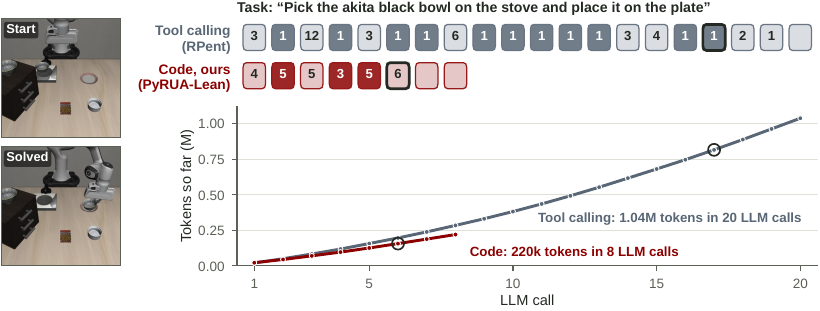}
  \caption{The episode of Figure~\ref{fig:method}, whole. \emph{Top:} each box is one LLM call; the number is how
    many tools (tool calling) or primitives (code) it called, dark boxes returned camera images, and the outlined box
    solved the task. \emph{Bottom:} token consumption so far; circles mark the LLM call that solved the task, reached after
    \V{DEMO_TOK_SOLVED_RPENT} tokens with tool calling and \V{DEMO_TOK_SOLVED_CODE} with code. \emph{Left:} the
    agent-view camera at the start and when the task was solved.}
  \label{fig:episode}
\end{figure}

Figure~\ref{fig:method} shows one stretch of an episode. Figure~\ref{fig:episode} and Table~\ref{tab:calls} follow
the whole episode with each agent: it is an instance both agents solved, close to the median cost of each on LIBERO-PRO
(spatial swap, task 7, seed 2). The tool-calling agent reads the guides, localizes, and then
moves one step per LLM call, getting three camera images back after every move; it solves the task in LLM call
\V{DEMO_SOLVED_RPENT}, after \V{DEMO_IMAGES_RPENT} images in all. The code agent localizes both objects in its first
cell, approaches and grasps in its second, closes the gripper and lifts the bowl in its third, checks the grasp in
its fourth, and carries and places the bowl in its fifth and sixth; it solves the task in LLM call \V{DEMO_SOLVED_CODE}
and asks for \V{DEMO_IMAGES_CODE} images in all. Appendix~\ref{app:episodes} follows further episodes in this way,
and Appendix~\ref{app:cells} describes what the code agent's cells do.

\begin{table}[h]
  \centering
  \caption{The two episodes of Figure~\ref{fig:episode}, LLM call by LLM call: what each call ran, and the token
    consumption so far. A check mark is the call that solved the task. Cached input is included, as everywhere.}
  \label{tab:calls}
  \scriptsize
  \setlength{\tabcolsep}{3pt}
  \begin{adjustbox}{max width=\linewidth}
  \begin{tabular}{@{}r>{\raggedright\arraybackslash}p{4.25cm}r>{\raggedright\arraybackslash}p{5.1cm}r@{}}
\toprule
Call & Tool calling (RPent): tools called & Tokens & Code (PyRUA-Lean): what the cell ran & Tokens \\
\midrule
1 & \texttt{read\_text\_file} $\times$3 & 20k & 4 primitive calls: \texttt{back\_project}, \texttt{segment}, \texttt{view\_env\_state} & 22k \\
2 & \texttt{view\_env\_state}; 3 images back & 50k & 5 primitive calls: \texttt{move\_to}, \texttt{pi0\_pick}, \texttt{view\_env\_state}; shows agentview and wrist & 44k \\
3 & \texttt{back\_project} $\times$12 & 83k & 5 primitive calls: \texttt{move\_to}, \texttt{segment}, \texttt{set\_gripper}, \texttt{view\_env\_state} & 70k \\
4 & \texttt{move\_to}; 3 images back & 118k & 3 primitive calls: \texttt{segment}; shows agentview and wrist & 96k \\
5 & \texttt{back\_project} $\times$3 & 157k & 5 primitive calls: \texttt{move\_to}, \texttt{segment}, \texttt{view\_env\_state}; shows agentview & 126k \\
6 & \texttt{move\_to}; 3 images back & 196k & 6 primitive calls: \texttt{move\_to}, \texttt{release}, \texttt{segment}, \texttt{view\_env\_state}\,$\checkmark$ & 156k \\
7 & \texttt{move\_to}; 3 images back & 239k & \texttt{finish} & 188k \\
8 & \texttt{back\_project} $\times$6 & 284k & final message & 220k \\
9 & \texttt{pi0\_pick}; 3 images back & 331k &  &  \\
10 & \texttt{set\_gripper}; 3 images back & 382k &  &  \\
11 & \texttt{move\_to}; 3 images back & 435k &  &  \\
12 & \texttt{move\_to}; 3 images back & 492k &  &  \\
13 & \texttt{move\_to}; 3 images back & 553k &  &  \\
14 & \texttt{back\_project} $\times$3 & 616k &  &  \\
15 & \texttt{back\_project} $\times$4 & 681k &  &  \\
16 & \texttt{move\_to}; 3 images back & 746k &  &  \\
17 & \texttt{release}; 3 images back\,$\checkmark$ & 816k &  &  \\
18 & \texttt{write\_text\_file} $\times$2 & 888k &  &  \\
19 & \texttt{finish} & 963k &  &  \\
20 & final message & 1.04M &  &  \\
\bottomrule
\end{tabular}

  \end{adjustbox}
\end{table}

\needspace{10\baselineskip}
Cells 1–6 of the code agent are reproduced exactly as generated by the model.
\begin{lstlisting}[title={\scriptsize LLM call 1}]
print(robo.state())
bowl = robo.segment('the black patterned bowl on the stove')
plate = robo.segment('the white plate with red rings')
print('bowl', bowl)
print('plate', plate)
print('table', robo.back_project(630, 610))
\end{lstlisting}
\Needspace{110pt}
\begin{lstlisting}[title={\scriptsize LLM call 2}]
assert bowl.found and plate.found
approach = robo.move_to([bowl.world_xyz[0], bowl.world_xyz[1], bowl.world_xyz[2]+0.19], gripper=-1)
print('approach', approach)
if approach.reached:
    pick = robo.pi0_pick('pick up the black bowl on the stove', max_chunks=20)
    print('pick', pick)
print(robo.state())
robo.show('agentview')
robo.show('wrist')
\end{lstlisting}
\Needspace{85pt}
\begin{lstlisting}[title={\scriptsize LLM call 3}]
print(robo.set_gripper(+1, steps=8))
held_state = robo.state()
print('lift', robo.move_to([held_state.eef_pos[0], held_state.eef_pos[1], 1.18], gripper=+1, step_clip=0.012))
held_bowl = robo.segment('the black patterned bowl held by the gripper')
print('held bowl', held_bowl)
print(robo.state())
\end{lstlisting}
\Needspace{60pt}
\begin{lstlisting}[title={\scriptsize LLM call 4}]
robo.show('agentview')
robo.show('wrist')
print('bowl now', robo.segment('the black patterned bowl on the stove'))
\end{lstlisting}
\Needspace{150pt}
\begin{lstlisting}[title={\scriptsize LLM call 5}]
held_bowl = robo.segment(point=(267, 339))
print('held localization', held_bowl)
assert held_bowl.found and held_bowl.world_xyz[2] > 1.05
carry_state = robo.state()
grip_offset = np.array(carry_state.eef_pos[:2]) - np.array(held_bowl.world_xyz[:2])
place_xy = np.array(plate.world_xyz[:2]) + grip_offset
print('grip offset', grip_offset, 'place xy', place_xy)
mid_xy = (np.array(carry_state.eef_pos[:2]) + place_xy)/2
for waypoint_xy in [mid_xy, place_xy]:
    move = robo.move_to([*waypoint_xy, carry_state.eef_pos[2]], gripper=+1, step_clip=0.012)
    print('carry', move)
    if not move.reached or move.terminated:
        break
robo.show('agentview')
\end{lstlisting}
\Needspace{200pt}
\begin{lstlisting}[title={\scriptsize LLM call 6}]
held_at_plate = robo.segment(point=(314, 795))
print('bowl above plate', held_at_plate)
assert held_at_plate.found
correction = np.array(plate.world_xyz[:2]) - np.array(held_at_plate.world_xyz[:2])
place_xy = np.array(robo.state().eef_pos[:2]) + correction
surface_map = robo.world_map()
bowl_patch = surface_map[290:365, 740:850]
bowl_heights = bowl_patch[:,:,2]
bowl_heights = bowl_heights[(bowl_heights>1.04) & (bowl_heights<1.17)]
bottom_z = float(np.quantile(bowl_heights, 0.02))
placement_z = plate.world_xyz[2] + robo.state().eef_pos[2] - bottom_z + 0.007
print('correction', correction, 'bottom', bottom_z, 'placement z', placement_z)
assert np.linalg.norm(correction)<0.07 and 0.95<placement_z<1.05
lower = robo.move_to([*place_xy, placement_z], gripper=+1, step_clip=0.012, tol=0.005, max_steps=100)
print('lower', lower)
if lower.reached and not robo.done:
    print('release', robo.release())
print('done', robo.done)
if not robo.done:
    robo.show('agentview')
\end{lstlisting}

\section{More episodes}
\label{app:episodes}

\begin{figure}[h]
  \centering
  \includegraphics[width=\linewidth]{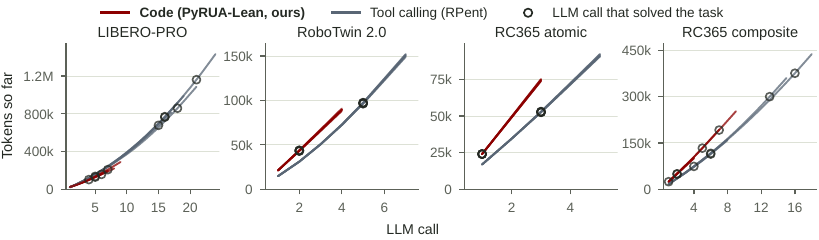}
  \caption{Token consumption so far of the episodes in Table~\ref{tab:episodes}, five per benchmark, LLM call by LLM
    call; circles mark the solving call. On RoboTwin~2.0 and RoboCasa365's atomic tasks the median episodes of both
    agents are nearly identical from task to task, so their lines overlap.}
  \label{fig:episodes}
\end{figure}

Table~\ref{tab:episodes} and Figure~\ref{fig:episodes} add five tasks per benchmark that both agents solved: on each
benchmark, the task instances whose token factor (tool calling's token consumption over code's, whole episodes) is
closest to the benchmark's median, at most one per task and none of Figure~\ref{fig:episode}'s task.

\begin{table}[h]
  \centering
  \caption{The episodes of Figure~\ref{fig:episodes}: LLM calls and token consumption of each whole episode, with
    tool calling (\emph{Tool}) and with code (\emph{Code}); shaded: how many times fewer tokens code consumed.}
  \label{tab:episodes}
  \footnotesize
  \setlength{\tabcolsep}{4pt}
  \begin{adjustbox}{max width=\linewidth}
  \begin{tabular}{@{}llc|cc|ccc@{}}
\toprule
 & & & \multicolumn{2}{c|}{LLM calls} & \multicolumn{3}{c}{Tokens} \\
Benchmark & Task & Seed & \hdtool & \hdcode & \hdtool & \hdcode & $\times$ \\
\midrule
LIBERO-PRO & \parbox[t]{4.4cm}{\raggedright object swap 0: Pick the alphabet soup and place it in the basket} & 1 & 21 & 8 & 1.09M & 219k & \gain{5.0$\times$} \\
 & \parbox[t]{4.4cm}{\raggedright spatial swap 5: Pick the akita black bowl on the ramekin and place it on the plate} & 1 & 18 & 7 & 918k & 188k & \gain{4.9$\times$} \\
 & \parbox[t]{4.4cm}{\raggedright goal swap 4: Put the bowl on the top of the drawer} & 3 & 19 & 7 & 991k & 199k & \gain{5.0$\times$} \\
 & \parbox[t]{4.4cm}{\raggedright spatial swap 9: Pick the akita black bowl on the wooden cabinet and place it on the plate} & 3 & 24 & 9 & 1.43M & 288k & \gain{5.0$\times$} \\
 & \parbox[t]{4.4cm}{\raggedright object swap 3: Pick the bbq sauce and place it in the basket} & 1 & 17 & 6 & 814k & 161k & \gain{5.0$\times$} \\
\midrule
RoboTwin 2.0 & \parbox[t]{4.4cm}{\raggedright place container plate} & 100000 & 7 & 4 & 151k & 89k & \gain{1.7$\times$} \\
 & \parbox[t]{4.4cm}{\raggedright open laptop} & 100008 & 7 & 4 & 149k & 88k & \gain{1.7$\times$} \\
 & \parbox[t]{4.4cm}{\raggedright place empty cup} & 100001 & 7 & 4 & 150k & 89k & \gain{1.7$\times$} \\
 & \parbox[t]{4.4cm}{\raggedright rotate qrcode} & 100001 & 7 & 4 & 151k & 89k & \gain{1.7$\times$} \\
 & \parbox[t]{4.4cm}{\raggedright move pillbottle pad} & 100002 & 7 & 4 & 152k & 91k & \gain{1.7$\times$} \\
\midrule
RoboCasa365 atomic & \parbox[t]{4.4cm}{\raggedright CloseToasterOvenDoor} & 1 & 5 & 3 & 91k & 74k & \gain{1.2$\times$} \\
 & \parbox[t]{4.4cm}{\raggedright PickPlaceToasterToCounter} & 4 & 5 & 3 & 92k & 75k & \gain{1.2$\times$} \\
 & \parbox[t]{4.4cm}{\raggedright CoffeeSetupMug} & 5 & 5 & 3 & 92k & 75k & \gain{1.2$\times$} \\
 & \parbox[t]{4.4cm}{\raggedright TurnOnSinkFaucet} & 4 & 5 & 3 & 91k & 74k & \gain{1.2$\times$} \\
 & \parbox[t]{4.4cm}{\raggedright TurnOnElectricKettle} & 4 & 5 & 3 & 91k & 75k & \gain{1.2$\times$} \\
\midrule
RoboCasa365 composite & \parbox[t]{4.4cm}{\raggedright ScrubCuttingBoard} & 3 & 8 & 4 & 159k & 101k & \gain{1.6$\times$} \\
 & \parbox[t]{4.4cm}{\raggedright WashLettuce} & 1 & 8 & 4 & 162k & 100k & \gain{1.6$\times$} \\
 & \parbox[t]{4.4cm}{\raggedright StackBowlsCabinet} & 1 & 6 & 3 & 117k & 78k & \gain{1.5$\times$} \\
 & \parbox[t]{4.4cm}{\raggedright CuttingToolSelection} & 2 & 18 & 9 & 438k & 253k & \gain{1.7$\times$} \\
 & \parbox[t]{4.4cm}{\raggedright LoadDishwasher} & 5 & 15 & 7 & 361k & 193k & \gain{1.9$\times$} \\
\bottomrule
\end{tabular}

  \end{adjustbox}
\end{table}

\section{What the code agent writes}
\label{app:cells}

We read the code agent's cells on every benchmark, in solved and failed episodes, with and without the VLA policy,
and then counted what they do over all \V{CODE_EPISODES_N} episodes of the main comparison, \V{CODE_CELLS_N} cells,
each episode cut at its budget (Table~\ref{tab:cells}).

\textbf{A cell acts more than a tool call does.} A code cell runs \V{ACT_PER_CELL} acting primitives on average
(moves, grasps, gripper commands and VLA calls), and \V{TWO_ACT_SHARE} of cells run two or more; an LLM call of tool
calling runs \V{ACT_PER_TOOLCALL}. Of tool calling's LLM calls that call a tool, \V{TOOL_ONE_SHARE} call exactly one;
of the \V{MULTI_CALLS} that call several, \V{MULTI_BP} batch back-projections and \V{MULTI_FILE} read or write
files, and only \V{ACTING_PAIRS} of all \V{TOOL_LLM_CALLS} run two acting tools together. Five habits of the code
agent make up the difference, and none of them fits in one tool call: the first two fold steps that tool calling
spreads over several LLM calls into one, the next two build what the tool list does not offer, and the last decides
when to look.

\textbf{Guarded chains} (\V{GUARDED_SHARE} of cells, \V{GUARDED_LIBERO} on LIBERO-PRO). A cell runs several
primitives in a row and makes each depend on the result of the one before (Listing~\ref{lst:guarded}): it moves
above the bowl, grasps with the VLA policy only if the move arrived and the task is not done, and looks only if it
is still not done. With tool calling, each link of such a chain is an LLM call.

\textbf{Retries} (\V{RETRY_SHARE} of cells, \V{RETRY_ROBOCASA} on RoboCasa365's atomic tasks). Where a VLA policy may
stop short of the goal, the cell runs it again until the task reports success (Listing~\ref{lst:retry}). With tool
calling, every retry is another LLM call, which brings back three camera images.

\textbf{Perception of its own} (\V{NUMPY_SHARE} of cells compute with NumPy, for perception or geometry). A cell can read a camera's image and point
cloud as arrays and locate objects itself: a color mask for a block, the points above a height for its top, the
principal axis of those points for its orientation, and from that the yaw of the grasp
(Listing~\ref{lst:perception}). A tool-calling agent gets what its tools compute (images, the 3D point at a pixel it
names, statistics of a box it names), and must work out anything else in its reasoning.

\textbf{Skills of its own} (\V{DEF_SHARE} of cells define a helper function, \V{HELPER_SHARE} call one that an
earlier cell defined). In Listing~\ref{lst:skill}, the agent writes a descent that moves down in 8\,mm steps and
stops as soon as a step does not reach its target, stops going down, or drifts sideways, and it reuses the function
in its next cell. Helpers like this are how the code agent grasps without a VLA policy (Section~\ref{sec:ablations}).

\textbf{Looking when needed} (\V{CONDLOOK_SHARE} of cells). The code agent looks often, \V{SHOW_SHARE} of its cells
ask for a camera image, but it chooses when and through which camera: \V{CONDLOOK_SHARE} of its cells look only if
the task is not yet done, where tool calling gets three images after every move whether it needs them or not.

\textbf{And single steps} (\V{SINGLE_STEP_SHARE} of cells, \V{SINGLE_STEP_ROBOTWIN} on RoboTwin~2.0). Many cells
take one step and look, like a tool call. On RoboTwin~2.0 the agent often runs the VLA policy one chunk at a time and
checks the result after each; code saves the fewest LLM calls there (Table~\ref{tab:main}).

\begin{table}[h]
  \centering
  \caption{What the code agent's cells do, over all \V{CODE_EPISODES_N} episodes of the main comparison (each cut at
    its budget), against what an LLM call of tool calling does on the same task instances. Acting primitives are
    moves, grasps, gripper commands and VLA calls. RC365: RoboCasa365.}
  \label{tab:cells}
  \footnotesize
  \begin{adjustbox}{max width=\linewidth}
  \begin{tabular}{@{}lccccc@{}}
\toprule
 & All & LIBERO-PRO & RoboTwin 2.0 & RC365 atomic & RC365 composite \\
\midrule
Acting primitives per code cell & 2.1 & 2.0 & 1.9 & 2.1 & 2.4 \\
Acting tools per LLM call of tool calling & 0.65 & 0.60 & 0.69 & 0.56 & 0.67 \\
\midrule
\multicolumn{6}{@{}l}{\emph{Share of code cells that}} \\
\quad run two or more acting primitives & 45\% & 61\% & 24\% & 50\% & 56\% \\
\quad take a single step and look & 35\% & 14\% & 54\% & 33\% & 28\% \\
\quad make an action depend on an earlier result & 19\% & 45\% & 12\% & 20\% & 14\% \\
\quad loop over motions & 14\% & 6\% & 12\% & 22\% & 17\% \\
\quad retry a VLA policy & 8\% & 0\% & 0\% & 18\% & 17\% \\
\quad compute with NumPy & 30\% & 25\% & 31\% & 29\% & 30\% \\
\quad reuse a variable from an earlier cell & 19\% & 33\% & 18\% & 10\% & 14\% \\
\quad call a helper function from an earlier cell & 7\% & 7\% & 6\% & 2\% & 9\% \\
\quad ask for a camera image & 84\% & 82\% & 78\% & 90\% & 90\% \\
\quad look only if the task is not done & 20\% & 19\% & 8\% & 58\% & 24\% \\
\bottomrule
\end{tabular}

  \end{adjustbox}
\end{table}

\begin{figure}[h]
\begin{lstlisting}
assert bowl.found and plate.found
bowl_xy = np.array(bowl.world_xyz[:2])
plate_xyz = np.array(plate.world_xyz)
table_z = robo.back_project(837,660).world_xyz[2]
prepose = [float(bowl_xy[0]),float(bowl_xy[1]),table_z+0.22]
approach = robo.move_to(prepose,gripper=robo.OPEN)
print('APPROACH',approach)
if approach.reached and not robo.done:
    picked = robo.pi0_pick('pick up the black bowl between the plate and the ramekin', max_chunks=20)
    print('PICK',picked)
print('STATE',robo.state())
if not robo.done:
    robo.show('agentview')
    robo.show('wrist')
\end{lstlisting}
\captionof{lstlisting}{A guarded chain: cell \V{EXAMPLE_GUARDED_CELL} of the code agent in
  \V{EXAMPLE_GUARDED_EPISODE}, exactly as the model wrote it.}
\label{lst:guarded}
\end{figure}

\begin{figure}[h]
\begin{lstlisting}
print(robo.task)
print(robo.success_criteria())
for attempt in range(3):
    result = robo.rldx_arm()
    print(result)
    print(robo.state())
    if robo.done or result.status != 'cap':
        break
if not robo.done:
    robo.show('agentview')
    robo.show('wrist')
\end{lstlisting}
\captionof{lstlisting}{A retry: cell \V{RETRY_CELL} of the code agent in \V{RETRY_EPISODE}, exactly as the model
  wrote it.}
\label{lst:retry}
\end{figure}

\begin{figure}[h]
\begin{lstlisting}
mask=red&(rows<92)&(cols>98)&(cols<165)&(world[...,2]>0.745)
points=world[mask]
upper=points[points[:,2]>0.794]
mean_xy=np.mean(upper[:,:2],axis=0)
eigenvalues,eigenvectors=np.linalg.eigh(np.cov(upper[:,:2].T))
long_axis=eigenvectors[:,1]
if long_axis[1]<0: long_axis=-long_axis
short_axis=np.array([long_axis[1],-long_axis[0]])
projection_long=upper[:,:2]@long_axis
projection_short=upper[:,:2]@short_axis
block_center=long_axis*((np.min(projection_long)+np.max(projection_long))/2)+short_axis*((np.min(projection_short)+np.max(projection_short))/2)
print('flat centre',block_center,'long axis',long_axis,'extents',np.ptp(projection_long),np.ptp(projection_short))
left_grasp_xy=block_center+0.045*long_axis
right_grasp_xy=block_center-0.05*long_axis
yaw=math.atan2(long_axis[1],long_axis[0])
q_down=np.array([math.cos(yaw/2)/math.sqrt(2),-math.sin(yaw/2)/math.sqrt(2),math.cos(yaw/2)/math.sqrt(2),math.sin(yaw/2)/math.sqrt(2)])
print('grasp',left_grasp_xy,'receive',right_grasp_xy,'quat',q_down)
safe=np.array(robo.state().left.eef_pos);safe[2]=1.03
if checked_move('left',safe,quat=robo.state().left.eef_quat,substeps=20):
    waypoint=np.array([left_grasp_xy[0],0.015,1.03])
    if checked_move('left',waypoint,quat=q_down,substeps=25):
        print(robo.move_to('left',[left_grasp_xy[0],left_grasp_xy[1],0.97],quat=q_down,substeps=25))
print(robo.state().left)
robo.show('head')
\end{lstlisting}
\captionof{lstlisting}{Perception of its own: cell \V{EXAMPLE_PERCEPTION_CELL} of the code agent in
  \V{EXAMPLE_PERCEPTION_EPISODE}, exactly as the model wrote it. \texttt{checked\_move}, a helper it defined in an
  earlier cell, moves an arm and reports whether the move arrived.}
\label{lst:perception}
\end{figure}

\begin{figure}[h]
\begin{lstlisting}
def lower_guarded(arm_name, target_tcp_z):
    initial=robo.state().arm(arm_name)
    target=np.array(initial.eef_pos)
    final_eef_z=target_tcp_z-0.12*initial.approach[2]
    while target[2]>final_eef_z+0.001 and not robo.done:
        before=robo.state().arm(arm_name)
        target[2]=max(final_eef_z,target[2]-0.008)
        result=robo.move_to(arm_name,target,substeps=8)
        after=robo.state().arm(arm_name)
        print('descent',result.reached,tuple(round(v,4) for v in after.tcp_pos))
        if result.terminated or not result.planned or not result.reached or after.eef_pos[2]>=before.eef_pos[2]-0.001 or np.linalg.norm(np.array(after.eef_pos[:2])-target[:2])>0.01:
            print('stop',result)
            break
lower_guarded('right',0.800)
robo.show('right_wrist')
robo.show('head')
\end{lstlisting}
\captionof{lstlisting}{A skill of its own: cell \V{EXAMPLE_SKILL_CELL} of the code agent in
  \V{EXAMPLE_SKILL_EPISODE}, exactly as the model wrote it.}
\label{lst:skill}
\end{figure}

\section{Instances only one agent solved}
\label{app:pairs}

\begin{figure}[p]
  \centering
  \includegraphics[width=\linewidth]{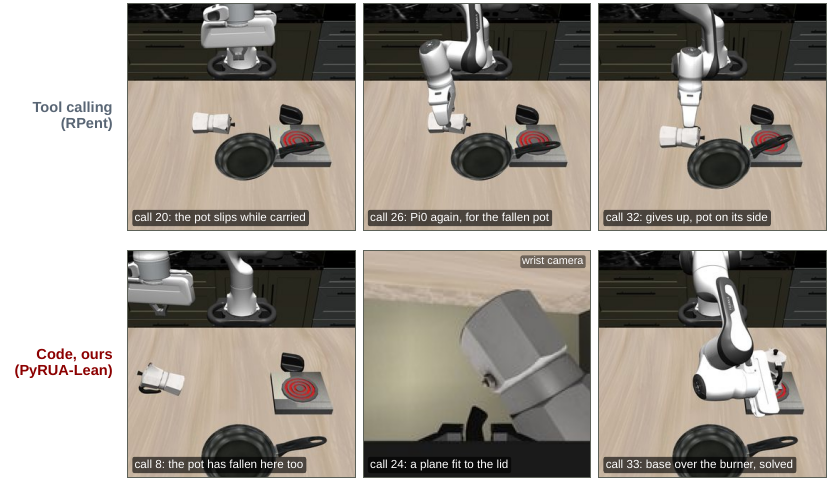}\\[5pt]
  \includegraphics[width=\linewidth]{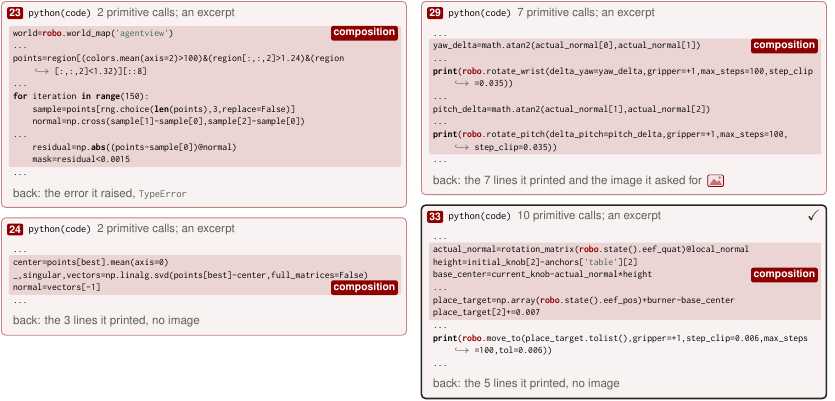}
  \caption{An instance only code solved: LIBERO-PRO, ``turn on the stove and put the moka pot on it'' (10 swap, task
    2, seed 3). Both agents turned the stove on and dropped the pot. Tool calling called the VLA pick primitive
    (\texttt{pi0\_pick}, which runs $\pi_{0.5}$; ``Pi0'' in the labels) on the fallen pot four times and ended the
    episode; the code agent grasped it itself, fitted a plane to the points of its lid to find how it
    leans, turned it upright, and set its base, not the gripper, over the burner. \emph{Top:} the recorded camera views
    after the LLM calls named. \emph{Bottom:} the code agent's cells, excerpts exactly as the model wrote them;
    ``\texttt{...}'' marks lines left out. The plane fit in LLM call 23 raised an error (NumPy's linear algebra
    refuses half-precision arrays); LLM call 24 cast the points and ran it again.}
  \label{fig:case-moka}
\end{figure}

\begin{figure}[p]
  \centering
  \includegraphics[width=\linewidth]{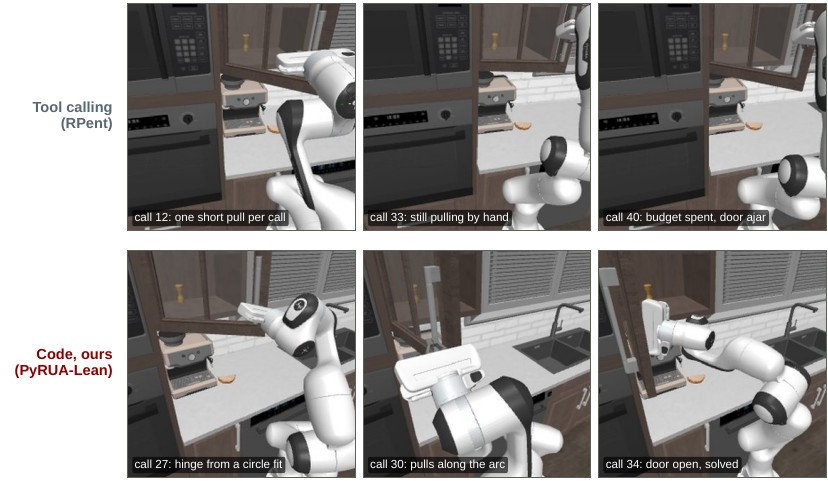}\\[5pt]
  \includegraphics[width=\linewidth]{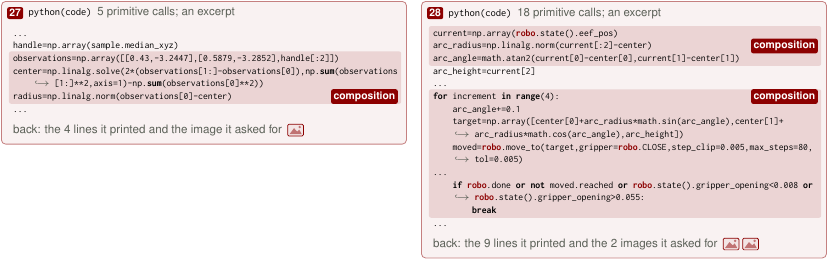}
  \caption{An instance only code solved: RoboCasa365 atomic, ``Open the cabinet door.'' (seed 4). Both agents
    located the door's hinge. Tool calling pulled the handle a few centimeters per LLM call and was still pulling when
    its 40 LLM calls ran out; the code agent took the hinge as the center of the circle through three positions of the
    handle, two of which it copied from earlier outputs, and pulled along that arc in one cell, stopping if a move fell
    short or the grip slipped. The door opened, and a VLA call in LLM call 34 completed the task. \emph{Top:} the recorded
    camera views; \emph{bottom:} the code agent's cells, excerpts exactly as the model wrote them.}
  \label{fig:case-cabinet}
\end{figure}

Code alone solves \V{CODE_ONLY_N} task instances of the main comparison, and tool calling alone \V{TOOL_ONLY_N}. In
\V{CODE_ONLY_CAP} of the \V{CODE_ONLY_N}, tool calling ran out of LLM calls; in the other \V{CODE_ONLY_ENDED}, it ended
the episode itself and reported failure. We read both episodes of each of those \V{CODE_ONLY_ENDED}. In about a quarter,
the code agent only called the VLA policy, and the policy succeeded in its episode but not in tool calling's. In most
of the others, both agents met the same setback, most often a VLA grasp that missed or an object that fell. Tool
calling then asked the policy again, or stopped with most of its budget left; the code agent closed the gripper itself,
scripted the recovery from the same primitives, checked each step, and in some episodes computed from the camera's
point cloud what no tool returns. Figures~\ref{fig:case-moka} and~\ref{fig:case-cabinet} follow two instances, one
where tool calling ended the episode itself and one where it ran out of LLM calls.

\section{Cost accounting}
\label{app:cost}

\textbf{Episode-level accounting.}
We removed memory-, recipe-, and audit-related instructions from the
prompts and operating guides, while retaining the baseline's original
tool descriptions. These descriptions may still prompt the baseline
to save episode-local recipe and audit files after success.
Such files are stored in the episode's own output directory, and
inspection of the evaluation traces found no cross-episode file reads.
Any associated post-success LLM calls are included in the reported
whole-episode metrics.

\textbf{Prices.} Every LLM call of every episode is billed from its own token usage at GPT-6 Astra's list prices as
of September 2026~\cite{openai2026pricing}: \$10 per million input tokens, \$1 for cached input, \$12.50 for cache
writes and \$50 for output, reasoning included. An LLM call above 272k input tokens would cost twice as much for input
and 1.5 times as much for output; none reached it.

\textbf{LLM calls, not seconds.} We count LLM calls rather than time episodes, because the latency of an LLM call on a
shared model gateway varies too much from one call to the next to compare.

\textbf{All-episode accounting.}
The efficiency metrics in Table~\ref{tab:main} are computed on
jointly solved instances, while success rates use all task instances.
Aggregating over all episodes, including failures, the
tool-calling-to-PyRUA-Lean input-token ratios are
\V{LIBERO_ALL_RATIO_1} on LIBERO-PRO,
\V{ROBOTWIN_ALL_RATIO_1} on RoboTwin~2.0,
\V{COMPOSITE_ALL_RATIO_1} on RoboCasa365 composite tasks, and
\V{ROBOCASA_ALL_RATIO_1} on RoboCasa365 atomic tasks;
the ratios of the estimated inference costs are
\V{S3_LIBERO_ALL_COST_RATIO}, \V{S3_ROBOTWIN_ALL_COST_RATIO}, \V{S3_COMPOSITE_ALL_COST_RATIO}, and \V{S3_ROBOCASA_ALL_COST_RATIO}.
On the atomic tasks an episode costs \$\V{S3_ROBOCASA_ALL_USD_TOOL} with tool calling and
\$\V{S3_ROBOCASA_ALL_USD_CODE} with PyRUA-Lean:
failed episodes on these tasks tend to approach the limit of 40 LLM calls
for both agents, while PyRUA-Lean uses more input tokens per LLM call.

\textbf{First prompts and output.} The median first prompt, which holds the system prompt, the guides and the tool
list or, for the code agent, the API reference and a task card with the first camera images, is \V{LIBERO_FIRST_RPENT} for tool calling and \V{LIBERO_FIRST_CODE} for code on LIBERO-PRO,
\V{ROBOTWIN_FIRST_RPENT} and \V{ROBOTWIN_FIRST_CODE} on RoboTwin~2.0, and \V{ROBOCASA_FIRST_RPENT} and
\V{ROBOCASA_FIRST_CODE} on RoboCasa365. On jointly solved instances, output tokens, reasoning included, are at most \V{OUT_SHARE_MAX} of either
agent's bill on any benchmark.

\section{Benchmarks and task instances}
\label{app:benchmarks}

Table~\ref{tab:benchmarks} lists what the main comparison ran: \V{TOTAL_MAIN_CELLS} task instances and
\V{TOTAL_MAIN_EPISODES} episodes. LIBERO-PRO runs every task at seeds 0--4, RoboTwin~2.0 at seeds 100000--100004
and RoboCasa365 at seeds 1--5. Where RoboTwin~2.0 cannot build a task at a seed, we use the next seed it can build,
and an episode that ended because its simulator or VLA server stopped answering was run again;
one RoboCasa365 instance (CoffeeSetupMug, seed 4) renders only noise on every camera, so that task runs at seed 6
instead.

\begin{table}[h]
  \centering
  \caption{The task instances of the main comparison. Each instance runs once with each agent.}
  \label{tab:benchmarks}
  \footnotesize
  \begin{tabular}{@{}lcccc@{}}
\toprule
Benchmark & Tasks & Seeds per task & Task instances & LLM-call budget \\
\midrule
LIBERO-PRO & 40 & 5 & 200 & 40 \\
RoboTwin 2.0 & 50 & 5 & 250 & 40 \\
RC365 atomic & 18 & 5 & 90 & 40 \\
RC365 composite & 32 & 5 & 160 & 100 \\
\midrule
All & 140 & & 700 & \\
\bottomrule
\end{tabular}

\end{table}

\section{Ablation settings}
\label{app:ablation}

The ablations of Table~\ref{tab:ablation} run on every task instance of the main comparison, at its seeds. Every
setting of a group uses the same task instances, and its first setting reuses the episodes of the main comparison.

\textbf{Tool calling with images on demand.} On every LIBERO-PRO task instance, the tool-calling agent ran once
more with one change: its tools return their text but no camera images, which it sees only when it calls RPent's
camera tool (three views per call), and its prompt no longer promises images after every tool call. Code needs no
such run, since a PyRUA-Lean cell returns images only when it asks for them; the Code column in Table~\ref{tab:observe} reuses the PyRUA-Lean episodes from the main comparison.
\begin{table}[h]
  \centering
  \caption{Tool calling with camera images after every move (the LIBERO-PRO baseline of Table~\ref{tab:ablation})
    and only on demand, against code, on all \V{OD_N} task instances of LIBERO-PRO. $^*$On the \V{OD_COMMON} instances
    all three solved (means).}
  \label{tab:observe}
  \footnotesize
  \begin{tabular}{@{}lccc@{}}
\toprule
 & \hdtool, images after every move & \hdtool, images on demand & \hdcode \\
\midrule
Success rate (\%), all task instances & 83.0 & 68.5 & 94.0 \\
Failures: gave up / LLM-call budget & 9 / 25 & 17 / 46 & 5 / 7 \\
LLM calls per solved episode$^*$ & 19.9 & 22.8 & 7.4 \\
Input tokens per solved episode$^*$ & 1.07M & 1.07M & 222k \\
\bottomrule
\end{tabular}

\end{table}

\section{RoboDojo: agent settings and checks}
\label{app:robodojo}

\textbf{Agent settings.} The settings are the same for all tasks and for both configurations: a budget of 100 LLM calls per episode, counted on
the calls that run a code cell, the task's own step limit, and a wall-clock limit of 90 minutes, which no episode
reached; images only when the code asks for them; at most 8,192
output tokens per LLM call, at high reasoning effort; no operating guide beyond the notes of
Section~\ref{sec:rd_system}; and a reminder, at most 20 times per episode, when an LLM call returns no code. One LLM
call returns at most one code cell: in the RGB configuration an episode runs \V{RD_CELLS_MEAN} cells in its \V{RD_LLM_MEAN} LLM calls on
average, and the remaining calls end the episode or return text only.

\textbf{Checks on the evaluation runs.} For all four seeds (one of the RGB configuration, three with depth), we checked that all 54
task names (a Generalization task and its randomized twin are two names) ran the protocol's number of episodes on
distinct layouts, that the three camera videos of every episode exist, that no episode ended on an error or a timeout
of the evaluation client, and that the verdict our adapter recorded agrees with the client's own result file for
every episode. The board-style values we compute from the episodes equal those of RoboDojo's own summary script.

\section{RoboDojo with depth: what differs}
\label{app:rd_depth}

The configuration of Section~\ref{sec:rd_depth} is the agent of Section~\ref{sec:robodojo} with one change to its
observation and to the three primitives that read it (Table~\ref{tab:rddiff}). The planner, the agent loop, the
budgets, the motion and gripper primitives, and the rest of the prompt are the same. Its success rate on each task is
the third column of Table~\ref{tab:rdtasks}. In this configuration an episode takes \V{RDD_LLM_MEAN} LLM calls on
average, and the three seeds cost about \$\V{RDD_USD_PROTOCOL_K},000 at our model gateway, or
\$\V{RDD_USD_EPISODE} per episode.

\begin{table}[h]
  \centering
  \caption{What differs between the two configurations of the RoboDojo agent.}
  \label{tab:rddiff}
  \footnotesize
  \setlength{\tabcolsep}{5pt}
  \begin{tabular}{@{}p{0.15\linewidth}p{0.39\linewidth}p{0.39\linewidth}@{}}
    \toprule
    & RGB (Section~\ref{sec:robodojo}) & RGB and depth (Section~\ref{sec:rd_depth}) \\
    \midrule
    Observation & Three RGB images and the robot state. & The same, plus
      a depth image and the intrinsic and extrinsic matrices of each camera, which are options of the client's
      observation configuration. \\
    \texttt{locate} & Takes the mask of the object in the camera asked, looks for the same object in the other two
      cameras, and reports the footprint and height that fit every camera that shows it. With one camera the height
      is a guess, and the result says so. & Takes the 3D points of the depth pixels under the mask, from one camera:
      they give center, top, and footprint directly. \\
    \texttt{back\_project} & Segments the object under the pixel and measures it as \texttt{locate} does; the answer
      is where the pixel's ray meets it, or the table. & Reads the depth at the pixel. \\
    \texttt{camera\_meta} & Worked out from the fixed calibration of the station and, for a wrist camera, from the
      arm's state. & The matrices the client sends. \\
    Operating notes & Say how to measure with the wrist cameras: what one camera cannot measure, and to look again
      from above before a grasp. & Not needed. \\
    \bottomrule
  \end{tabular}
\end{table}

\section{PyRUA-Lean Coder: training settings}
\label{app:coder}

Table~\ref{tab:coderhp} lists the settings with which PyRUA-Lean Coder v0.1-4B was trained
(Section~\ref{sec:coder_training}).

\begin{table}[h]
  \centering
  \caption{Training settings of PyRUA-Lean Coder v0.1-4B.}
  \label{tab:coderhp}
  \footnotesize
  \begin{tabular}{@{}ll@{}}
    \toprule
    Base model & Qwen3.5-4B, fine-tuned in full \\
    Algorithm & GRPO; clip range 0.2 below, 0.28 above; no KL term, no entropy term \\
    Reward & 1 if the benchmark's success check passes, 0 otherwise \\
    Training instances & 45: LIBERO-PRO goal, spatial and object suites, 3 tasks $\times$ 5 seeds each \\
    Episodes per iteration & 16 instances $\times$ 8 rollouts $=$ 128 \\
    Optimizer steps per iteration & 2, of 64 episodes each \\
    Optimizer & Adam, learning rate $10^{-6}$ (constant), $\beta=(0.9, 0.98)$, weight decay 0.1 \\
    Iterations & 30 (3,840 episodes, 60 optimizer steps) \\
    Rollout sampling & temperature 1.0 \\
    Limits of a training episode & 24 code cells, 30 LLM calls, 25 minutes; 4,096 new tokens per LLM call \\
    Generated tokens per episode & at most 40,960; context 65,536 tokens \\
    Hardware and software & one node with eight GPUs; miles for training, SGLang for rollouts; about twelve hours \\
    \bottomrule
  \end{tabular}
\end{table}

\clearpage
\section{RoboDojo: results per task}
\label{app:rd_tasks}

Table~\ref{tab:rdtasks} lists the \V{RD_TASKS_N} tasks of RoboDojo with the success rate of PyRUA-Lean, of PyRUA-Lean
with depth (Section~\ref{sec:rd_depth}), of GPT-6 Astra controlling the end-effector directly, and of the best entry of
the public board on that task. PyRUA-Lean is above every entry on \V{RD_TASK_WINS} tasks and level with the best on
\V{RD_TASK_TIES}.

\begin{table}[!h]
  \centering
  \caption{Success rate (\%) per RoboDojo task: PyRUA-Lean (\V{RD_SEED_PHRASE}), PyRUA-Lean with depth
    (\V{RDD_SEED_PHRASE}; Section~\ref{sec:rd_depth}, not compared with the board), GPT-6 Astra with direct
    end-effector control~\cite{zhang2026unexpected} (\emph{Astra}), and the best entry of the public board dated
    \V{RD_BOARD_DATE} on that task (\emph{Best}, among the 57 policies with per-task results; where several hold it, their number is given, and a dash where it is 0.0). A Generalization task's
    value pools its standard and randomized episodes. Bold: PyRUA-Lean above every entry of the board.}
  \label{tab:rdtasks}
  \scriptsize
  \setlength{\tabcolsep}{5pt}
  \begin{tabular}{@{}lrr|rrl@{}}
\toprule
Task & Ours & \makecell[r]{Ours,\\with depth} & Astra & Best & Best board entry \\
\midrule
\multicolumn{6}{@{}l}{\emph{Generalization} (12 tasks)} \\
\texttt{stack\_bowls} & {\bfseries 78.0} & 75.3 & 62.0 & 74.7 & Simate-beta \\
\texttt{push\_T} & {\bfseries 76.0} & 78.0 & 60.0 & 60.0 & GPT-6 Astra (direct) \\
\texttt{pack\_objects\_into\_box} & {\bfseries 32.0} & 17.3 & 0.0 & 17.3 & Simate-beta \\
\texttt{fold\_clothes} & 70.0 & 48.7 & 72.0 & 78.0 & Simate-beta \\
\texttt{hang\_mugs} & 2.0 & 14.0 & 0.0 & 32.0 & VPP2 \\
\texttt{sweep\_blocks} & {\bfseries 18.0} & 24.7 & 4.0 & 14.0 & Rex-M1 Preview \\
\texttt{pour\_liquid\_into\_cup} & 8.0 & 16.0 & 4.0 & 67.3 & Rex-M1 Preview \\
\texttt{make\_toast} & 6.0 & 18.7 & 0.0 & 39.3 & VPP2 \\
\texttt{arrange\_largest\_number} & {\bfseries 86.0} & 76.7 & 60.0 & 60.0 & GPT-6 Astra (direct) \\
\texttt{sort\_nesting\_dolls\_by\_size} & {\bfseries 74.0} & 76.0 & 18.0 & 18.7 & VPP2 \\
\texttt{store\_laptop\_and\_headphones} & 0.0 & 0.0 & 0.0 & 34.0 & Simate-beta \\
\texttt{stack\_blocks} & {\bfseries 100.0} & 100.0 & 86.0 & 86.0 & GPT-6 Astra (direct) \\
\midrule
\multicolumn{6}{@{}l}{\emph{Precision} (8 tasks)} \\
\texttt{fasten\_screws} & {\bfseries 66.0} & 69.3 & 16.0 & 48.0 & Awomo-0.5 \\
\texttt{plug\_in\_charger} & 12.0 & 14.0 & 0.0 & 30.7 & PhysicalRSI \\
\texttt{insert\_tubes} & 44.0 & 70.7 & 0.0 & 74.7 & VPP2 \\
\texttt{pour\_balls\_into\_vase} & {\bfseries 76.0} & 69.3 & 4.0 & 60.7 & VPP2 \\
\texttt{play\_Xylophone} & {\bfseries 64.0} & 67.3 & 8.0 & 30.7 & Awomo-0.5 \\
\texttt{deposit\_coin} & 34.0 & 34.0 & 2.0 & 56.0 & PhysicalRSI \\
\texttt{insert\_key} & 0.0 & 0.0 & 0.0 & 39.3 & PhysicalRSI \\
\texttt{build\_tower} & 2.0 & 4.0 & 2.0 & 83.3 & Simate-beta \\
\midrule
\multicolumn{6}{@{}l}{\emph{Long-Horizon} (8 tasks)} \\
\texttt{put\_bottles\_into\_dustbin} & 76.0 & 94.7 & 10.0 & 100.0 & Simate-beta \\
\texttt{fill\_pen\_holder} & 14.0 & 16.7 & 0.0 & 42.0 & InternW0-$\Delta$ \\
\texttt{classify\_objects} & {\bfseries 98.0} & 99.3 & 56.0 & 56.0 & GPT-6 Astra (direct) \\
\texttt{play\_tic\_tac\_toe} & 94.0 & 94.7 & 0.0 & 94.0 & PhysicalRSI \\
\texttt{fill\_egg\_holder} & {\bfseries 2.0} & 2.0 & 0.0 & 1.0 & Liber-0 Lite \\
\texttt{organize\_table} & {\bfseries 34.0} & 39.3 & 0.0 & 26.0 & Simate-beta \\
\texttt{make\_kong} & 0.0 & 0.0 & 0.0 & 90.0 & GalaxeaVLA (G0.5) \\
\texttt{play\_stacking\_toy} & {\bfseries 66.0} & 67.3 & 0.0 & 23.3 & VPP2 \\
\midrule
\multicolumn{6}{@{}l}{\emph{Memory} (6 tasks)} \\
\texttt{cover\_blocks} & 100.0 & 100.0 & 46.0 & 100.0 & 8 entries \\
\texttt{match\_and\_pick\_from\_conveyor} & 82.0 & 86.7 & 62.0 & 92.0 & InternW0-$\Delta$ \\
\texttt{swap\_blocks} & 0.0 & 0.0 & 0.0 & 38.7 & Simate-beta \\
\texttt{swap\_T} & {\bfseries 98.0} & 100.0 & 18.0 & 92.7 & PhysicalRSI \\
\texttt{press\_by\_number} & 0.0 & 0.0 & 70.0 & 96.0 & VPP2 \\
\texttt{imitate\_sorting\_sequence} & {\bfseries 98.0} & 97.3 & 36.0 & 36.0 & GPT-6 Astra (direct) \\
\midrule
\multicolumn{6}{@{}l}{\emph{Open} (8 tasks)} \\
\texttt{align\_blocks} & 78.0 & 71.3 & 50.0 & 86.0 & PhysicalRSI \\
\texttt{general\_pickup} & {\bfseries 98.0} & 93.3 & 84.0 & 84.0 & GPT-6 Astra (direct) \\
\texttt{stack\_blocks\_by\_language} & {\bfseries 100.0} & 100.0 & 40.0 & 40.0 & GPT-6 Astra (direct) \\
\texttt{solve\_equation} & {\bfseries 90.0} & 86.7 & 40.0 & 60.0 & PhysicalRSI \\
\texttt{classify\_objects\_by\_language} & {\bfseries 92.0} & 90.7 & 30.0 & 30.0 & GPT-6 Astra (direct) \\
\texttt{pick\_from\_conveyor\_by\_image} & {\bfseries 34.0} & 28.7 & 4.0 & 4.0 & GPT-6 Astra (direct) \\
\texttt{store\_tools\_in\_toolbox} & 0.0 & 0.0 & 0.0 & 0.0 & -- \\
\texttt{pour\_by\_language} & {\bfseries 20.0} & 27.3 & 0.0 & 0.0 & -- \\
\bottomrule
\end{tabular}

\end{table}

\clearpage
\ifdraftmarks
\noindent\fcolorbox{pendc}{white}{\parbox{\dimexpr\linewidth-2\fboxsep-2\fboxrule}{\small
  \textcolor{pendc}{\textbf{Draft status of v2 (remove with \texttt{\textbackslash draftmarksfalse}).}}
  Text and numbers in \pend{this colour} are provisional: they come from runs that are not finished and are filled
  in by \texttt{scripts/v2\_numbers.py} when the result files change; \pend{--} is a value that does not exist yet.
  \begin{itemize}
\item nothing: every run this report quotes is finished
\end{itemize}}}
\fi

\end{document}